\documentclass[10pt,journal,compsoc]{IEEEtran}
  \usepackage[nocompress]{cite}

  \usepackage[pdftex]{graphicx}
  \graphicspath{{./images/}}
  \DeclareGraphicsExtensions{.pdf,.jpg,.jpeg,.png,.tif}
  \usepackage[caption=false,font=footnotesize,labelfont=sf,textfont=sf]{subfig}

\usepackage{amsmath}
\usepackage{amsfonts}
\usepackage{dsfont}
\usepackage{array}

\usepackage{fixltx2e}
\usepackage{dblfloatfix}

\usepackage{url}
\usepackage{lipsum}
\usepackage{xcolor,colortbl,moreverb}
\usepackage{color}
\usepackage[hidelinks,bookmarks=true]{hyperref}
\usepackage{bookmark}
\usepackage[normalem]{ulem}
\usepackage{pdfpages}
\usepackage{glossaries}
\usepackage{footnote}

\usepackage{xargs}

\newcommandx{\chris}[1]{{\color{blue} \textit{[chke]} #1}}
\newcommandx{\ullu}[1]{{\color{purple} \textit{[ullu]} #1}}
\newcommandx{\wail}[1]{{\color{cyan} \textit{[wamus]} #1}}
\newcommandx{\abda}[1]{{\color{violet} \textit{[abda]} #1}}
\newcommandx{\jake}[1]{{\color{yellow} \textit{[jake]} #1}}
\newcommandx{\strike}[1]{{\color{red} \sout{#1}}}
\newcommandx{\todo}[1]{{\color{orange} \textbf{TODO:} #1}}
\newcommandx{\blue}[1]{{\color{blue} #1}}

\makeglossaries

\newacronym{CT}{CT}{computed tomography}
\newacronym{MRI}{MRI}{Magnet Resonance Imaging}
\newacronym{DTI}{DTI}{Diffusion Tensor Imaging}
\newacronym{DEM}{DEM}{Digital Elevation Model}
\newacronym{LiDAR}{LiDAR}{Light Detection and Range}
\newacronym[\glsshortpluralkey={LoD's},\glslongpluralkey={Levels-of-Detail}]{LoD}{LoD}{Level-of-Detail}
\newacronym{KML}{KML}{Keyhole Markup Language}
\newacronym{PID}{PID}{Proportional-Integral-Differential}
\newacronym{PBR}{PBR}{Point-based Rendering}
\newacronym{SVM}{SVM}{Support Vector Machine}
\newacronym{RLE}{RLE}{Runlength Encoding}
\newacronym{VDB}{VDB}{Volumetric Dynamic Grid B+Tree}
\newacronym[\glsshortpluralkey={LoA's},\glslongpluralkey={Levels-of-Abstraction}]{LoA}{LoA}{Level-of-Abstraction}
\newacronym[\glsshortpluralkey={GPUs},\glslongpluralkey={Graphics Processing Units}]{GPU}{GPU}{Graphics Processing Unit}
\newacronym{ISO}{ISO}{International Standardization Organization}
\newacronym[\glsshortpluralkey={IDs},\glslongpluralkey={Identifiers}]{ID}{ID}{Identifier}
\newacronym[\glsshortpluralkey={SDIs},\glslongpluralkey={Spatial Data Infrastructures}]{SDI}{SDI}{Spatial Data Infrastructure}
\newacronym[\glsshortpluralkey={TINs},\glslongpluralkey={Triangulated Irregular Networks}]{TIN}{TIN}{Triangulated Irregular Network}
\newacronym{GML}{GML}{Geography Markup Language}
\newacronym{XML}{XML}{Extensible Markup Language}
\newacronym{VRML}{VRML}{Virtual Reality Markup Language}
\newacronym[\glsshortpluralkey={GIS},\glslongpluralkey={Geographic Information Systems}]{GIS}{GIS}{Geographic Information System}
\newacronym[\glsshortpluralkey={DBMS},\glslongpluralkey={Database Management Systems}]{DBMS}{DBMS}{Database Management System}
\newacronym{OGR}{OGR}{OGR Simple Features Library}
\newacronym{GDAL}{GDAL}{Geospatial Data Abstraction Library}
\newacronym{VOG}{VOG}{Virtual Outcrop Geology}
\newacronym{SBA}{SBA}{Sparse Bundle Adjustment}
\newacronym{MPS}{MPS}{Multiple-Point Statistics}
\newacronym{DLT}{DLT}{Direct Linear Transform}
\newacronym{MPCD}{MPCD}{Mobile Personal Communication Device}
\newacronym{MI}{MI}{Mutual Information}
\newacronym{SLAM}{SLAM}{Simultaneous Localisation and Mapping}
\newacronym{SIFT}{SIFT}{Scale-Invariant Feature Transform}
\newacronym{SURF}{SURF}{Speeded-Up Robust Features}
\newacronym{SfM}{SfM}{structure from motion}
\newacronym{RANSAC}{RANSAC}{Random Sampling Consensus}
\newacronym{EPnP}{EPnP}{Efficient Perspective-n-Point}
\newacronym{ICP}{ICP}{Iterative Closest Point}
\newacronym{VGI}{VGI}{Volunteered Geographic Information}
\newacronym{UAV}{UAV}{Unmanned Aerial Vehicle}
\newacronym{TLS}{TLS}{terrestrial laser scanning}
\newacronym{TI}{TI}{training image}
\newacronym{SNR}{SNR}{signal-to-noise ratio}
\newacronym{FBP}{FBP}{filtered back-projection}
\newacronym{ART-TV}{ART-TV}{algebraic reconstruction technique with total variation}
\newacronym[\glsshortpluralkey={LAGs},\glslongpluralkey={liquids, aerosols and gels}]{LAG}{LAG}{liquid, aerosol and gel}
\newacronym{EM}{EM}{electromagnetic}
\newacronym{E-M}{E-M}{expectation maximisation}
\newacronym{CIL}{CIL}{checked-in luggage}
\newacronym{FAMS}{FAMS}{fast adaptive mean shift}
\newacronym{MAR}{MAR}{metal artefact reduction}
\newacronym{MS}{MS}{mean shift}
\newacronym{MST}{MST}{minimum spanning tree}
\newacronym{MECT}{MECT}{multi energy CT}
\newacronym{DECT}{DECT}{dual energy CT}
\newacronym{SECT}{SECT}{single energy CT}
\newacronym{DEI}{DEI}{dual energy index}
\newacronym{SVD}{SVD}{singular value decomposition}
\newacronym{LDA}{LDA}{linear discriminant analysis}
\newacronym{SDA}{SDA}{sparse discriminant analysis}
\newacronym{CNN}{CNN}{convolutional neural network}
\newacronym{kNN}{kNN}{k-nearest neighbour}
\newacronym{MCMC}{MCMC}{monte carlo markov chain}
\newacronym{LAC}{LAC}{linear attenuation coefficient}
\newacronym{CNR}{CNR}{contrast-to-noise ratio}
\newacronym{MITK}{MITK}{the Medical imaging Interaction Ioolkit}
\newacronym{DeVIDE}{DeVIDE}{the Delft Visualisation and Image processing Development Environment}
\newacronym{DVR}{DVR}{direct volume rendering}
\newacronym{LSH}{LSH}{locally sensitive hashing}
\newacronym{GMM}{GMM}{gaussian mixture models}
\newacronym{MRF}{MRF}{markov random field}
\newacronym{ROI}{ROI}{region-of-interest}
\newacronym{CCD}{CCD}{charge-coupled device}

\begin{document}
%
\title{Multi-Spectral Imaging via Computed Tomography (MUSIC) - Comparing Unsupervised Spectral Segmentations for Material Differentiation}

%

\author{Christian~Kehl,
        Wail~Mustafa,
        Jan~Kehres,
        Ulrik~Lund~Olsen,
        and~Anders~Bjorholm~Dahl
\IEEEcompsocitemizethanks{\IEEEcompsocthanksitem C. Kehl, W. Mustafa and A.B. Dahl are with the Department
of Computing (DTU Compute), Technical University of Denmark, 2800 Kongens Lyngby, Denmark.\protect\\
E-mail: \{chke,wamus,abda\}@dtu.dk
\IEEEcompsocthanksitem J. Kehres and U.L. Olsen are with the Department
of Physics (DTU Physics), Technical University of Denmark, 2800 Kongens Lyngby, Denmark.\protect\\E-mail: \{jake,ullu\}@dtu.dk}
\thanks{Manuscript received XXXX YY, ZZZZ; revised XXXX YY, ZZZZ.}}

%
%

\ifCLASSOPTIONpeerreview
  \markboth{Transaction on Image Processing,~Vol.~XX, No.~YY, MONTH~2018}%
{AUTH_1 \MakeLowercase{\textit{et al.}}: Multi-Spectral Imaging via Computed Tomography}
\else
  \markboth{Transaction on Image Processing,~Vol.~XX, No.~YY, MONTH~2018}%
{Kehl \MakeLowercase{\textit{et al.}}: Multi-Spectral Imaging via Computed Tomography}
\fi

\IEEEtitleabstractindextext{%
\begin{abstract}
Multi-spectral computed tomography is an emerging technology for the non-destructive identification of object materials and the study of their physical properties. Applications of this technology can be found in various scientific and industrial contexts, such as luggage scanning at airports. Material distinction and its identification is challenging, even with spectral x-ray information, due to acquisition noise, tomographic reconstruction artefacts and scanning setup application constraints. We present MUSIC -- and open access multi-spectral CT dataset in 2D and 3D -- to promote further research in the area of material identification. We demonstrate the value of this dataset on the image analysis challenge of object segmentation purely based on the spectral response of its composing materials. In this context, we compare the segmentation accuracy of fast adaptive mean shift (FAMS) and unconstrained graph cuts on both datasets. We further discuss the impact of reconstruction artefacts and segmentation controls on the achievable results. Dataset, related software packages and further documentation are made available to the imaging community in an open-access manner to promote further data-driven research on the subject.
\end{abstract}

}

\maketitle

\IEEEdisplaynontitleabstractindextext

%
\IEEEpeerreviewmaketitle

\IEEEraisesectionheading{\section{Introduction}\label{sec:introduction}}

\begin{figure}[htbp]
\centering
\subfloat{\includegraphics[width=0.249\linewidth,height=4.1cm,keepaspectratio]{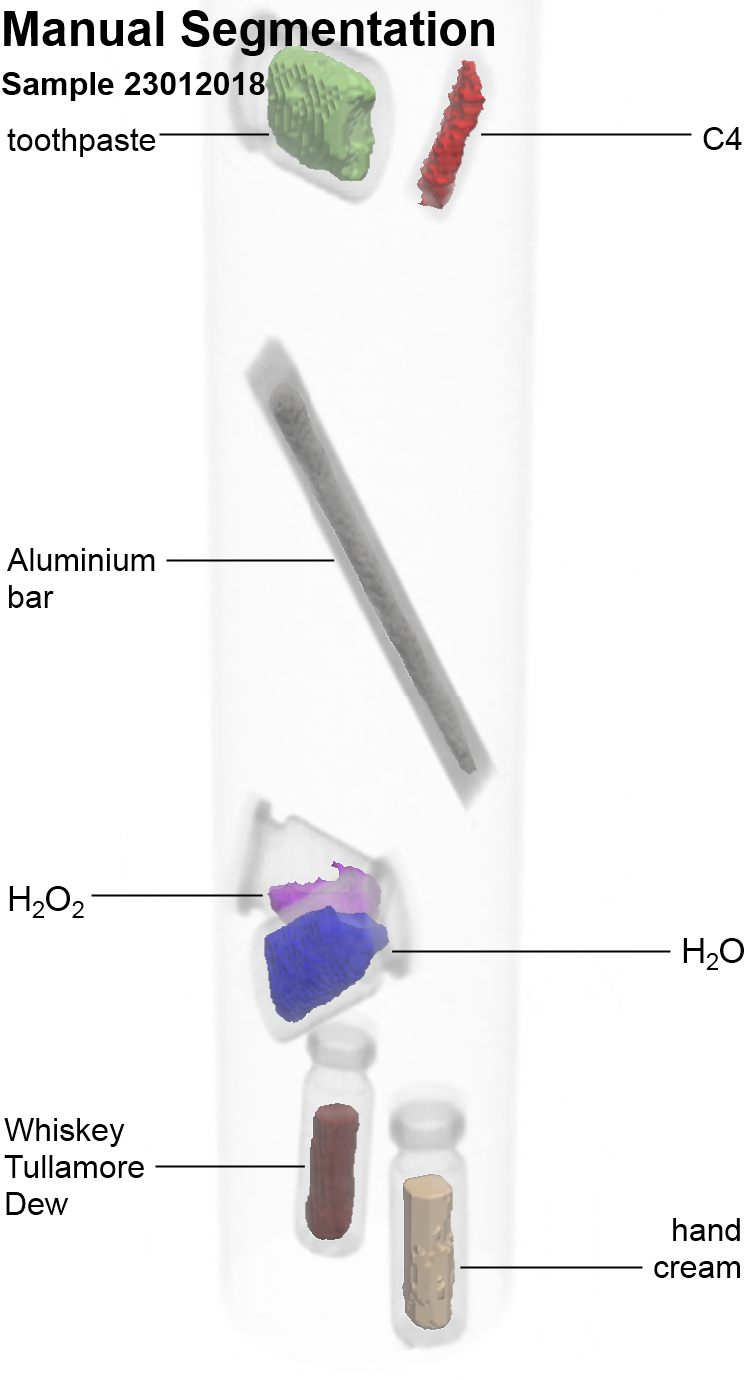}%
\label{fig:segmentationResultPreview:manualSegmentation}}
\subfloat{\includegraphics[width=0.249\linewidth,height=4.1cm,keepaspectratio]{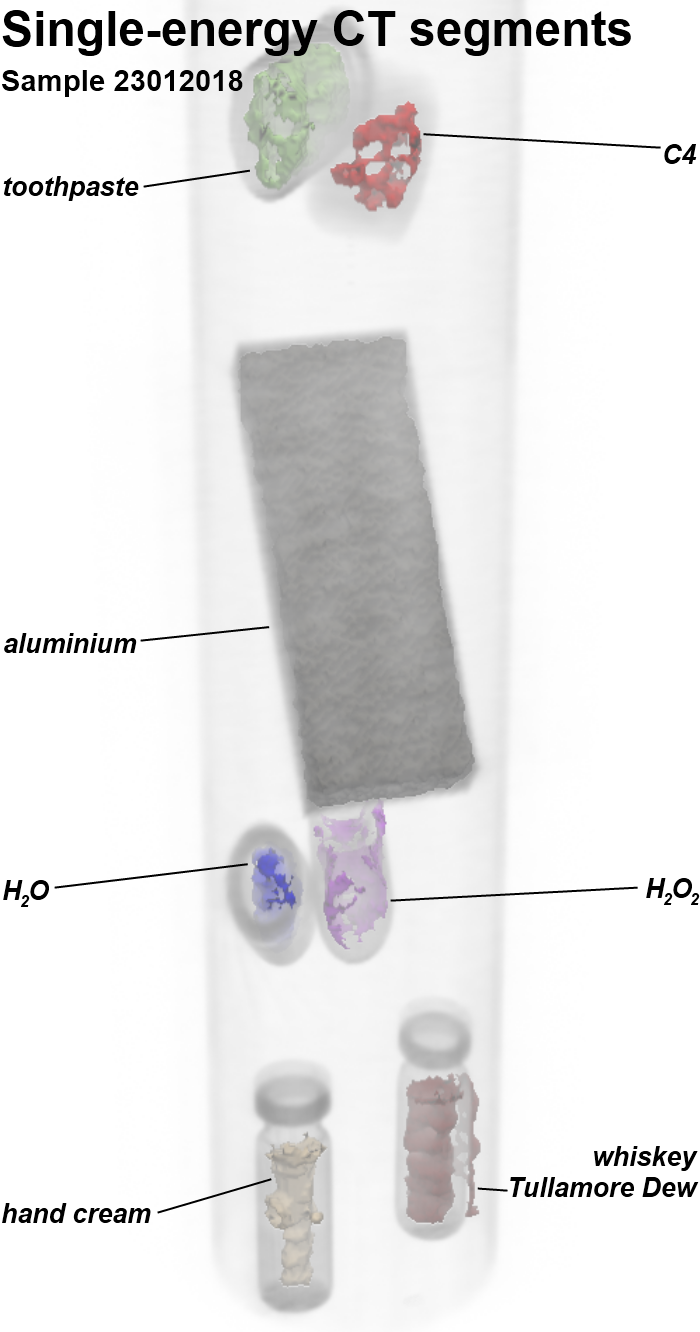}%
\label{fig:segmentationResultPreview:SEsegmentation}}
\subfloat{\includegraphics[width=0.249\linewidth,height=4.1cm,keepaspectratio]{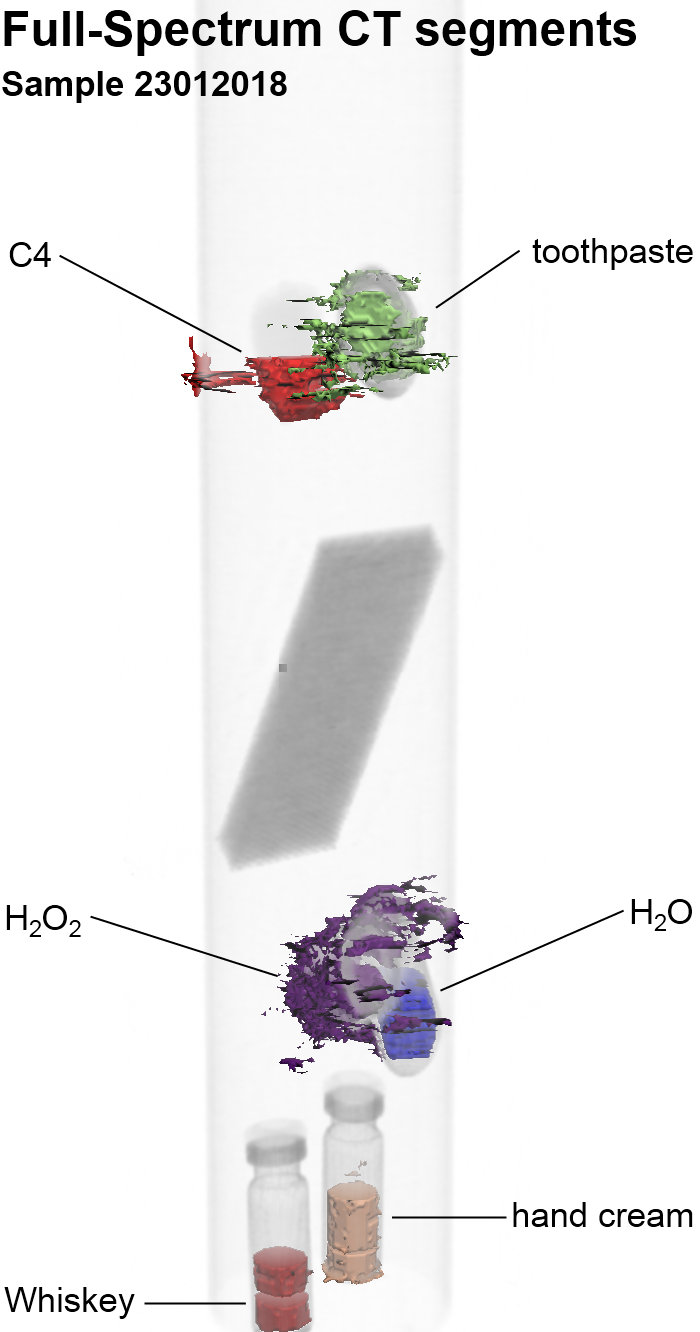}%
\label{fig:segmentationResultPreview:fullSpectrumSegmentation}}
\subfloat{\includegraphics[width=0.249\linewidth,height=4.1cm,keepaspectratio]{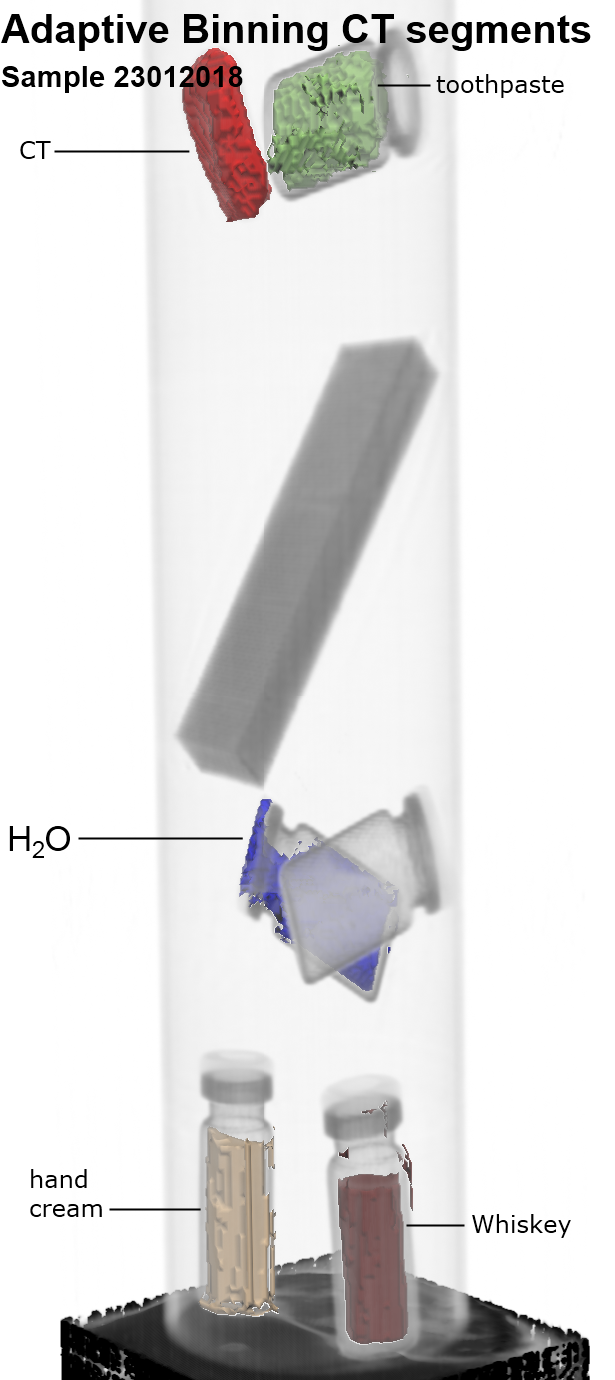}%
\label{fig:segmentationResultPreview:adaptBinnedSegmentation}}
\caption{The added value of using spectral CT data for fluid segmentation: a single-energy CT (b) segmentation suffers from lots of noise and the inability to clearly differentiate between materials, compared to manual reference segmentations (a). A full-spectrum analysis can differentiate between the materials, while still suffering in quality from low signal-to-noise ratios (c). The proposed method (d) is best able to differentiate individual segments that are closest to given manual reference outlines.}
\label{fig:segmentationResultPreview}
\end{figure}

\IEEEPARstart{S}{pectral} \gls{CT} is researched in physics applications with the potential to be "the emerging technology" for material distinction of objects\cite{Fornaro2011,McCollough2015}. Due to advances in flux handling of the detector technology made in recent years, \gls{MECT} instruments are becoming a feasible alternative to their single- and dual energy counterparts. Deducing materials with anisotropic spectral properties within volumetric, tomographic scans is more involved than the spectral analysis of natural \cite{Jordan2013}- or hyperspectral remote sensing \cite{Chang2003} images and usually requires a set of prior assumptions on object behaviour or prior knowledge on image content. In this work, we present an approach for deducing object boundaries, represented by segmentation masks, based on material properties in spectral volume \gls{CT} scans, covering large parts of the x-ray spectrum. The presented approach is superior to previous segmentations of \gls{SECT} data with established methods (such as Otsu's method \cite{Otsu1979} or domain-specific tools \cite{MarkReport}), which is demonstrated in fig. \ref{fig:segmentationResultPreview}. The spectral dataset, which is made available to the community, also provides input for training supervised segmentation methods, e.g. via neural networks. The material differentiation objective itself is synonymous with object segmentation based on material properties. The computed segmentation maps provide an input to improve subsequently material identification (i.e. object classification).

Material differentiation via spectral imaging has various application scenarios related to the imaging modality, ranging from well-known medical diagnosis and monitoring \cite{McCollough2015} over manufacture quality control \cite{Einarsson2017,Einarsdottir2014} to geological evaluation in the planetary sciences and for natural resources. The focus of this article is material differentiation in tomographic reconstructions, for which we extend previously published segmentation methods, originally presented in other contexts and on other imaging modalities. The target application of our research is in reliable fluid threat item detection (e.g. explosives, acids) within the \gls{CIL} scanning project\footnotemark[1] for airport security. Despite the strong focus on spectral tomography, the presented segmentation method extensions also apply to other sources of spectral image data.

\begin{figure*}[!htbp]
\centering
\includegraphics[width=0.98\textwidth]{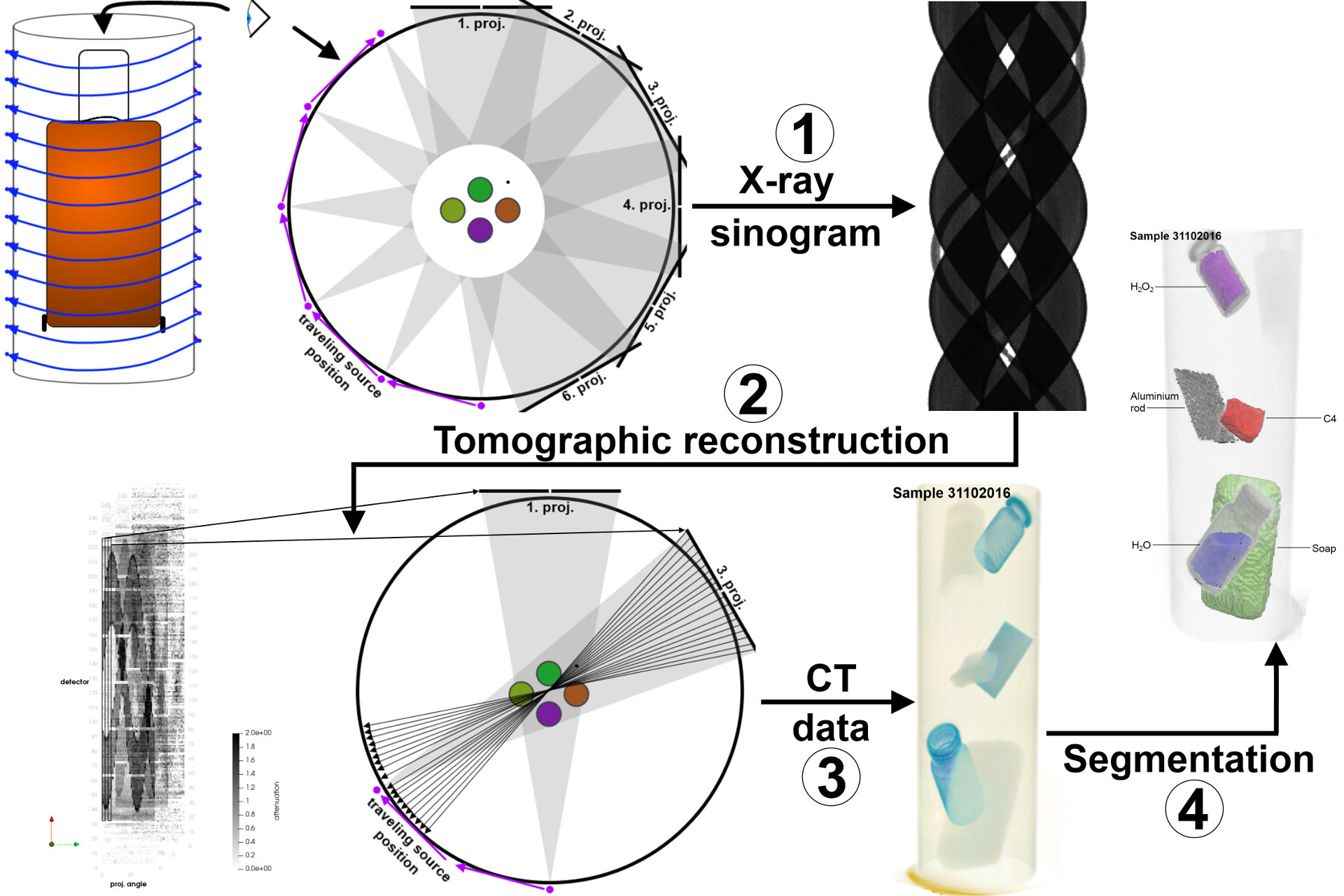}%
\caption{\label{fig:banner} Sketch of the tomographic data processing for segmenting MECT data in the CIL project.}
\end{figure*}

\begin{figure}[tbh]
\begin{center}
  \centering
  \includegraphics[width=0.98\linewidth]{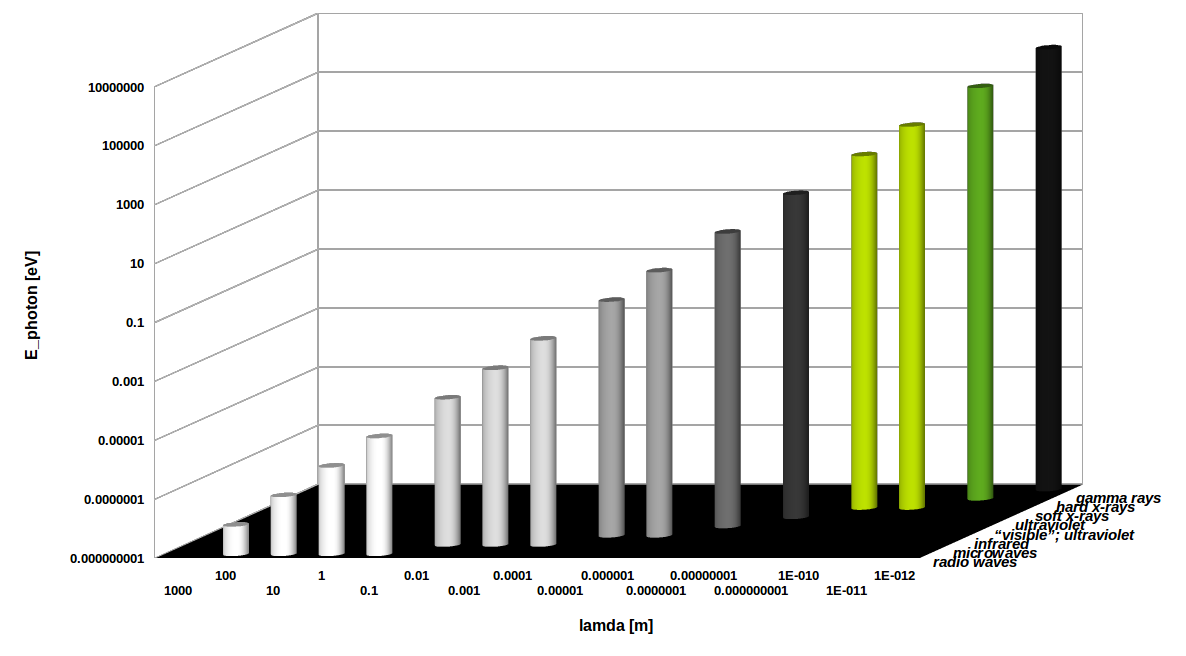}
  \caption{Chart of the electromagnetic spectrum. The range of x-ray spectrum used in this work is highlighted in green.}
  \label{fig:intro:emspectrum}
\end{center}
\end{figure}

A sketch of the tomographic data processing is shown in fig.\ref{fig:banner}. In contrast to other application areas where the range of imaged materials is limited and objects are easier to separate, luggage scanning deals with a wide range of material compositions, tightly arranged next to one another and inadequately distinguished by just the material phase (e.g. fluid, viscous semi-fluid, solid lattice with air pockets, solid filled, metal). Therefore, information from multiple parts of the x-ray spectrum are needed to distinguish objects and materials in more detail (e.g. different types and concentrations of fluids, see fig. \ref{fig:intro:emspectrum} for reference of the used spectrum). Moreover, as fluids are non-rigid and arbitrarily arranged in luggage, an actual tomographic reconstruction is required to detect their traces. Luggage scanning is therefore an ideal testcase for spectral segmentation.

\footnotetext[1]{CIL2018 NextGen Scanner for Checked In Luggage - \url{https://innovationsfonden.dk/en/node/786}}

In terms of computational requirements, we distinguish between three image segmentation- and classification scenarios: (i) supervised, where the expected content is known in advance or can be compared to a reference database of known materials and segmentation maps (e.g. medical diagnosis and monitoring, atlas-based segmentation\cite{Dogdas2007}); (ii) semi-supervised, where external annotations and input can be used to create a base segmentation and classification (e.g. quality control); (iii) unsupervised, where no prior knowledge about the acquired content is available. Methodologically, semi-supervised and unsupervised methods differ in that the number of separate objects and materials can be determined and indicated in semi-supervised schemes, while this information is commonly unknown for unsupervised schemes. this passage is more a discussion for classification - as we don't do classification here, leave that one out.

Spectral (i.e. multivariate) multi-label object distinction is a long-standing challenge in image analysis and various algorithms have been proposed to address this task. A short overview of existing methods applicable to \textbf{spectral} \gls{CT} segmentation is given in section \ref{sec:literature}. Utilizing recent advances of neural network-based image segmentation \cite{LeCun2015} requires extensive, domain-specific training datasets. Hence, one major gap addressed by this article is the lack of an openly available benchmark dataset for spectral CT segmentation in order to test the plethora of proposed techniques in a standardized manner. We introduce the spectral 2D (i.e. MUSIC2D) and 3D (i.e. MUSIC3D) \gls{CT} datasets for algorithmic benchmarking of multivariate image segmentation methods. Both datasets are made publicly available with manual segmentations for reference- and scoring purpose. Apart from spectral tomographic segmentation, the dataset also provides input for research in CT correction algorithms, such as \gls{MAR}. The benchmark dataset is curated collaboratively by physicists and computer scientists at the Technical University of Denmark (DTU) for future assessments. Based on benchmark segmentation results, we discuss the effects of reconstruction artefacts and spectral binning on the visual quality and quantitative precision of the detected materials and objects.

%
%

\section{Scientific background and related literature}\label{sec:literature}

The following paragraphs give an overview of tasks inherent to tomographic segmentation and the proposed approaches within the literature to address them. The radiographic scanning applied here follows a common workflow for \gls{CT} acquisition and processing (see fig.\ref{fig:banner}): the x-ray acquisition system collects spectral radiographic projections of the object of interest in a given scanning setup. Correction algorithms are applied to account for detector response and photon correlation ambiguities (see Dreier et al. \cite{Dreier2018} for details). Then, a sinogram for each energy bin is computed. The sinograms are used in an iterative tomographic reconstruction algorithm \cite{Sidky2008} to compute the interior x-ray attenuation contribution (i.e. the \glspl{LAC}) per energy bin. Algorithms such as \gls{MAR} and spectral range reduction can be employed beforehand or be integrated in reconstruction process to improve the smoothness and reduce noise in the tomographic dataset. The improved reconstruction data are then subject to the object segmentation algorithms assessed in this article. Due to over-segmentation effects, a fusion of segments based on the statistical distribution or morphological constraints may be required for each material. This latter task is performed manually in the remainder of the article for reasons of brevity and due to the focus on global segmentation methods.


\subsection{CT reconstruction}\label{sec:literature:reconstruction}

Common fan-beam \gls{CT} reconstruction aims at computing the x-ray attenuation in a 2D image from multiple 1D angular radiographic projections (analogously: computing the 3D volume from 2D angular projections) \cite{Kak1988}. The resulting \gls{LAC} of a scanned object depends on its material properties (e.g. density and element composition). 

In general, the \gls{CT} reconstruction task is an inverse problem \cite{Hansen2010}. A reconstruction can be computed analytically using the inverse Radon transform (i.e. \gls{FBP}; \cite{Bracewell1967,Kak1988}), if sufficient projections for a given image resolution are available. The acquisition of such excessive number of projections for high-resolution \gls{CT} images is infeasible in some practical scenarios due to (i) physical constraints of the imaging system , (ii) computational runtime demands or (iii) x-ray dose of radiation-sensitive objects (e.g. medical applications). On the other hand, an acquisition with too few projections lowers the \gls{SNR} and introduces currently-unrecoverable reconstruction artefacts in the tomographic images.

Compressed sensing theory- and reconstruction methods \cite{Donoho2006} circumvent the excessive x-ray acquisition while still preserving a high quality in the image reconstruction. Total variation methods introduced geometric constraints and regularisation terms to solve inverse imaging more adequately \cite{Hansen2010}. Within our experiments, we use an \gls{ART-TV} for the \gls{CT} reconstruction \cite{Sidky2006,Sidky2008}. Still, by using as few as 9 up to 37 projections with 256 detector pixels each for reconstructing slices of 100x100 pixels in resolution, the \gls{SNR} poses a major challenge to unsupervised image analysis approaches.

\subsection{Unsupervised tomographic image segmentation}\label{sec:literature:imgseg}
For the purpose of this article, we define the image segmentation $S_T$, result of an iterative process over $t \in [0;T]$, as extracting non-overlapping regions $s \in S$ whose pixel intensities $f(x) \ \forall x \in X$ are maximally homogeneous whereas their average intensity are maximally heterogeneous to all other regions. Conversely, we can also state the problem as minimising the inter-segment similarity while maximising the in-segment similarity (see eq. \ref{segm:similarity:eq1} to \ref{segm:similarity:eq4}).

\begin{equation}
\label{segm:similarity:eq1}
sim_i = \sqrt{\sum_{x=0}^{N} (f(x) - \mathrm{f(x)})}
\end{equation}
\begin{equation}
\label{segm:similarity:eq2}
s_i = s \  \forall s \in S_t, t= [0;T]
\end{equation}
\begin{equation}
\label{segm:similarity:eq3}
sim(s_i,s_j) = 1 - \frac{|\mathrm{f(x | X \in s_i)} - \mathrm{f(x | X \in s_j)}|}{\mathrm{f(x)}_{max}}
\end{equation}
\begin{eqnarray}
\label{segm:similarity:eq4}
S_T = \text{min} \ sim(s_i, s_j | \text{max} \  (sim_i) \  \cap \text{max} \  (sim_j)) \\
\forall (s_i,s_j) \in S, s_i \neq s_j
\end{eqnarray}

\begin{figure}[htbp]
\centering
\subfloat{\includegraphics[width=0.98\linewidth]{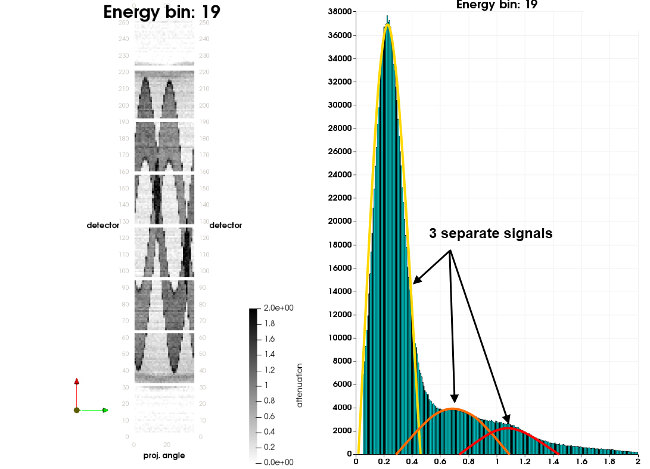}%
\label{fig:spectralSeg:projections:e19}}
\hfil
\subfloat{\includegraphics[width=0.98\linewidth]{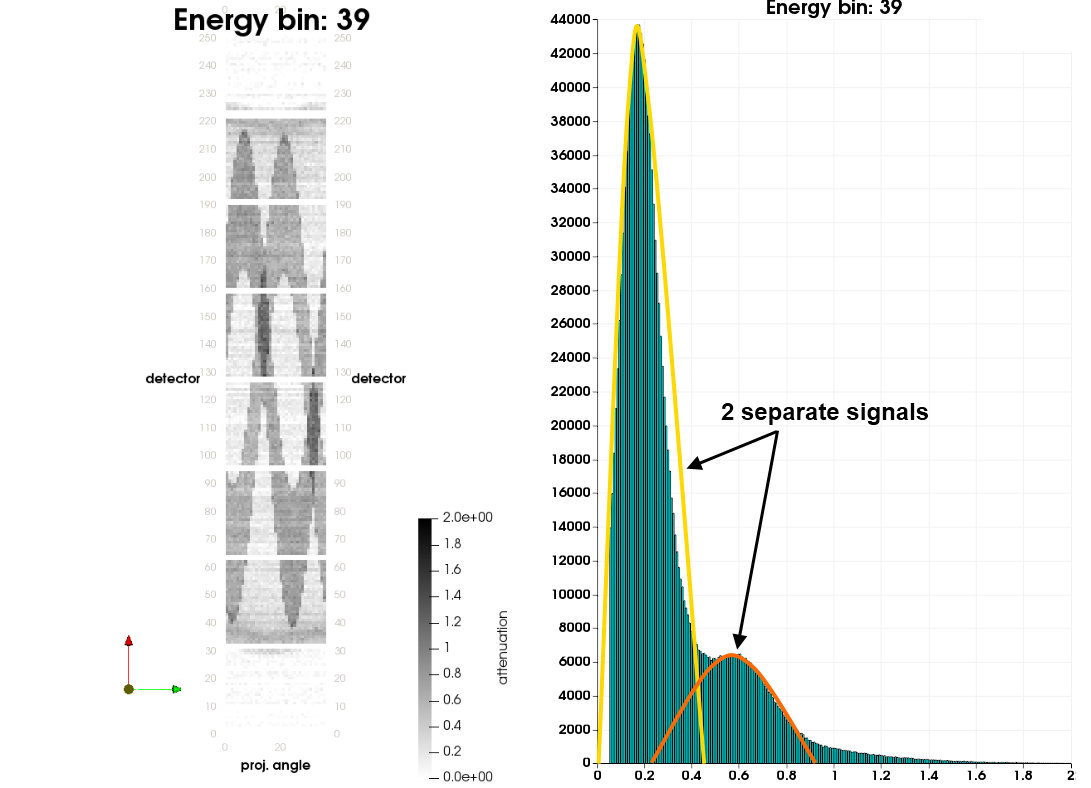}%
\label{fig:spectralSeg:projections:e39}}
\hfil
\subfloat{\includegraphics[width=0.98\linewidth]{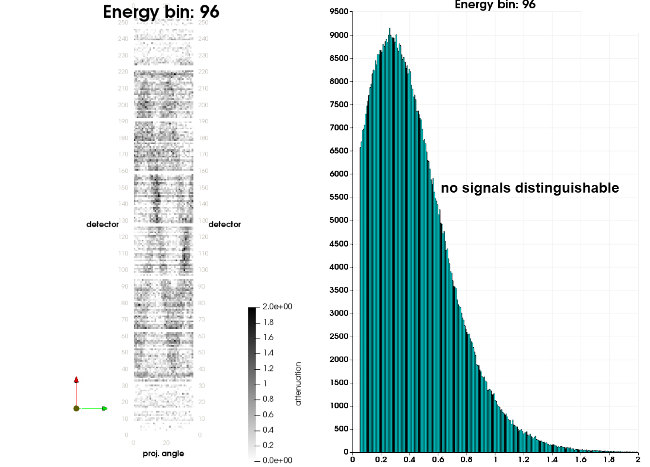}%
\label{fig:spectralSeg:projections:e96}}
\caption{Side-by-side comparison of the x-ray projections and their statistical intensity distribution (using a histogram) for energy bins 19 (a), 39 (b) and 96 (c), corresponding to energy levels 40.52 keV, 92.60 keV and 120.72 keV.}
\label{fig:spectralSeg:projections}
\end{figure}

\begin{figure}[htbp]
\centering
\subfloat{\includegraphics[width=0.98\linewidth]{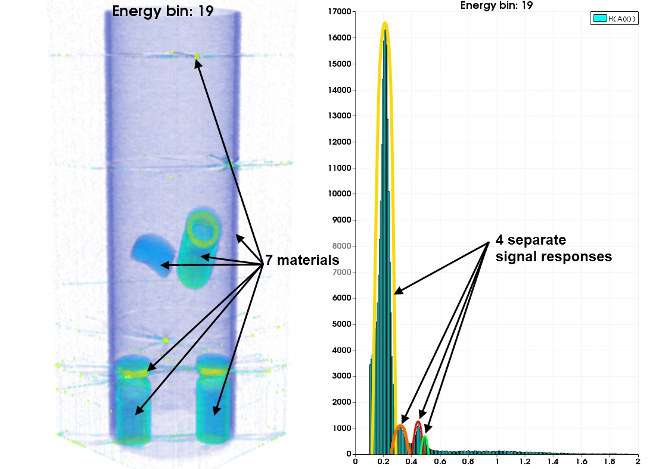}%
\label{fig:spectralSeg:reconstructions:e19}}
\hfil
\subfloat{\includegraphics[width=0.98\linewidth]{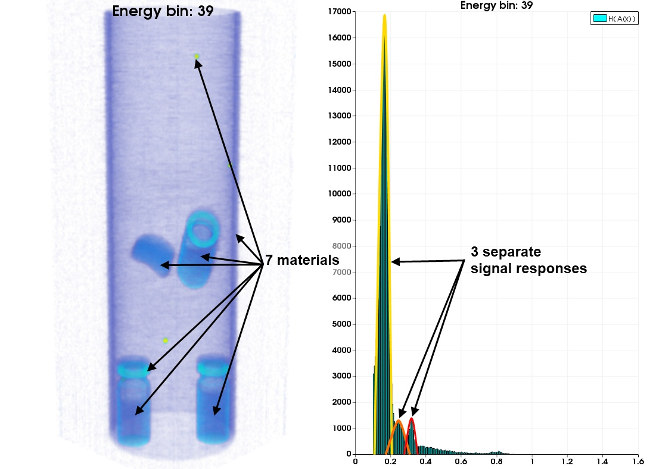}%
\label{fig:spectralSeg:reconstructions:e39}}
\hfil
\subfloat{\includegraphics[width=0.98\linewidth]{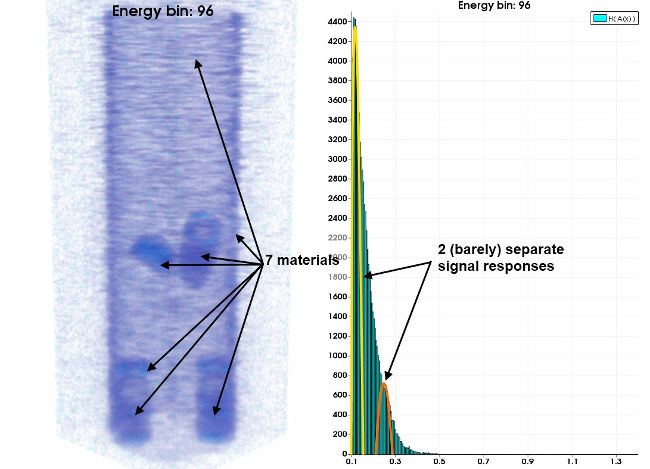}%
\label{fig:spectralSeg:reconstructions:e96}}
\caption{Side-by-side comparison of the tomographic reconstructions and their statistical LAC distribution (see histogram) for energy bins 19 (a), 39 (b) and 96 (c), corresponding to energy levels 40.52 keV, 92.60 keV and 120.72 keV.}
\label{fig:spectralSeg:reconstructions}
\end{figure}

When segmenting spectral data, previous approaches \cite{Einarsdottir2014,Einarsson2017} utilise statistical distributions and \textit{a priori} knowledge (e.g. number of expected materials and objects) as conditioning parameters to simplify the segmentation task. These information are hard to obtain in a general acquisition setup: The statistical distribution of x-ray intensities (fig. \ref{fig:spectralSeg:projections}) and material attenuations (fig. \ref{fig:spectralSeg:reconstructions}) does not allow for a clear separation and derivation of the number of scanned materials. As visible in the related images, the segmentation task can even be perceptually challenging for humans.

This work focusses on the fully automatic, unsupervised segmentation of 2D- and 3D spectral CT images. Therefore, an in-depth discussion on conditioned graph cut methods \cite{Boykov2004,Rother2004,Boykov2006} or active contour models \cite{Kass1988,Sapiro1997,Xu1998} is not within the scope of this article, other than being applicable to the image segmentation if the number of distinct materials or objects and their approximate position were known \textit{a priori}. Comparable approaches that either need prior conditioning (\gls{GMM}\cite{Einarsdottir2014}, \gls{SDA} \cite{Clemmensen2011,Einarsson2017}) or that are related to supervised segmentation, such as random walks \cite{Tu2002} via texture synthesis or \gls{SVM} learning \cite{CampsValls2014}, are described elsewhere as they are not directly applicable to the given problem.

Traditional multi-label unsupervised image segmentation techniques without prior knowledge of the image content include approaches such as hierarchical clustering \cite{Wu1993} and \gls{MST} \cite{Banerjee2014}. More recently, \gls{MS} algorithms \cite{Comaniciu2002} have been used in multivariate segmentation \cite{Tao2007}, which shares similarities to spectral segmentation. The drawback of fixed kernel sizes for analysing anisotropic value distributions has been addressed by adaptive \gls{MS} \cite{Georgescu2003,Shimshoni2006}. In this work, we utilise the \gls{FAMS} algorithm \cite{Georgescu2003} together with its spectral gradient extension \cite{Jordan2013} as one reference procedure for unbiased, unconstrained spectral CT segmentation in 2D and 3D. 

Directly mapping algorithms for analysing multivariate statistics to image analysis challenges (e.g. using \gls{MS} for image segmentation) tends to omit the dense connectivity between data points inherent in images. The Power Watershed algorithm by Couprie et al. represents a recent approach for unsupervised image segmentation that utilizes dense connectivity constraints \cite{Couprie2011}, which has not been tested in this paper due to the computational complexity of the watershed calculation across multiple channels. In this work, we extend an unconditioned graph cut algorithm by Felzenszwalb and Huttenlocher \cite{Felzenszwalb2004} to perform the image segmentation while taking advantage of the inherent dense pixel connectivity. The specific graph cut formulation utilised in this article is explained in section \ref{sec:algorithms:graphCut}.

%
%
%

\subsection{Object- and material identification in luggage scanning}\label{sec:literature:luggageScanning}

In the baggage screening application, we first distinguish between the object identification and the material identification, which also approximately relates to the distinction between the segmentation and classification. Separating and locating objects within the \gls{CT} scan is a segmentation challenge. Identifying rigid threat items, such as small arms, has been the identification focus in traditional literature (e.g. \cite{Singh2004,Abidi2006}), which requires a good object segmentation combined with simple shape priors in the classification.

Recent interest has shifted to the identification of explosives and \gls{LAG} threats in baggage. These objects are non rigid and occur in arbitrary shapes. In such cases, the segmentation is only an auxiliary tool to group areas of similar x-ray intensity or attenuation. In turn, the utilization of attenuation contributions within the volume necessitates an actual tomographic reconstruction whereas rigid threat item detection can equally be performed on the radiographic projections themselves.

Classic single-energy \gls{CT}- and x-ray radiography is still prevalently used for the identification of rigid threat items due to the rapid acquisition and processing. Furthermore, for solid objects, the captured attenuation relates well to the material density and the rough distinction between solids (e.g. plastics and metals). Liquid materials generally appear as low-attenuation objects in single-energy CT and different liquids exhibit almost identical attenuation values, which is why baggage screening for such threats uses \gls{DECT} \cite{Heitz2010} or \gls{MECT} \cite{Eger2011a,Megherbi2013}. The \gls{DEI} is used in \gls{DECT} to distinct materials on a chemical level \cite{Mouton2015b}. The use of \gls{MECT} is a promising recent research direction due to its additional material information, though \gls{CT} reconstruction and image processing algorithms are more complex. An review of the state-of-the-art in baggage screening is provided by Mouton et al. \cite{Mouton2015a}, which is based on the final project report of the ALERT initiative and supplementary review extensions.

More specific to \gls{DECT} and \gls{MECT} segmentation, Eger et al. attempt to use the full energy spectrum (10keV to 150keV) for material identification via extensive chemical reference samples and direct, pixelwise \gls{LAC} comparison \cite{Eger2011a}. Then, they perform a linear dimensionality reduction via \gls{SVD} and \gls{LDA} to extract the most relevant image information, while the final classification is performed via trained classifiers (i.e. likelihood ratio test) for reference material data. Mouton et al. provide a benchmark overview of four chained \gls{DECT} segmentation methods, consisting of reference material comparison via \gls{DEI}, a subsequent quality estimation and the refinement of critical segment areas via connected components and split-and-merge strategies \cite{Mouton2015b}. The presented results are acceptable on visual inspection, though note that the evaluation dataset depicts mainly rigid objects and easily distinguishable materials. Martin et al. presented a learning-based segmentation algorithm for \gls{DECT} data that trains a probabilistic \gls{kNN} classifier while employing unconditioned graph cuts for the class separation \cite{Martin2015}. Our approach of using the spectral information as explicit auxiliary dimension is similar to \cite{Martin2015}, though we explicitly address the segmentation without prior knowledge on the number of objects (e.g. the knowledge of \textit{k} for \gls{kNN}).

Currently published material classification methods that purely operate on a material database comparison achieve limited success due to reconstruction artefacts, noise and natural material variations, which is discussed in the literature \cite{Eger2011a,Mouton2015b,Martin2015}. More recently, material classification is approached with pattern recognition and machine learning algorithms such as visual bag-of-words \cite{Bastan2011}, with a classification accuracy of 70\% (see review in \cite{Kundegorski2016}). A recent trend in material classification is the training of \glspl{CNN} to perform object segmentation and material classification. Mery et al. first attempted to use pre-trained models on the ImageNet dataset from the two prevalent neural network architectures with mixed success. Akcay et al. use a pre-trained model with ImageNet data from ConvNet \cite{Akcay2016}, which is refined using actual 2D reconstructions of a limited-size dataset from the UK airport authorities \cite{Akcay2018}. The final-stage classification uses \glspl{SVM}. Both research groups report the lack of sufficient training data as a major impediment for better detection accuracy with \glspl{CNN}. 

Possibly one of the largest application areas of the the 2D and 3D spectral dataset published with this paper is the provision of an overall set of 2376 spectral images from the domain of material science that can be directly used as input for advanced machine learning algorithms (e.g. \glspl{CNN}, training-based \gls{MCMC} \cite{Strebelle2001,Tu2002}) in baggage screening.

\section{Acquisition Instrumentation and Parameters}\label{sec:instruments}

The 2D- and 3D spectral datasets in this article are acquired with a custom build tomography setup using MULTIX ME-100 V2 cadmium telluride (CdTe) x-ray detectors. 
It counts  photons in the energy range of 20keV to 160keV and resolves them in upto 128 energy bins \cite{Kehres2017,Olsen2017}. The detector itself consists of two elements of each 128 pixels.  The x-ray generator is a microfocus  Hamamatsu ML12161-07. For the datasets presented here the operation Voltage was 150 kVp and 250µA resulting in a focal spot size of ~75µm. The beam is collimated horizontally with a JJ X-ray IB-80-Air to a height of 0.6mm at the source a custom built 5mm thick tungsten slit directly in front of the detector with an opening of 0.6mm  keeps scattered photons from the detector. 

A spectral correction algorithm is applied to the real data in order to remove spectral detector pixel bevels artefacts \cite{Dreier2018}. The result are 37 energy-corrected x-ray projections covering the full angular range of 360 degrees. These spectral sinograms are included with each dataset. Subsequently, we use an \gls{ART-TV} reconstruction algorithm to obtain the spectral images and volumes. The lateral reconstruction pixel resolution is limited by the amount of projections and the detector pixel resolution, result currently in a target resolution of 100 x 100 pixels laterally (i.e. per slice). The datasets are still subject to metal artefacts as proper \gls{MAR} is still under development in our acquisition procedure.


\section{The Datasets}

In this section we present the datasets (2D- and 3D-spectral) that form the major contribution of this article. We discuss the core properties of each dataset, such as \gls{SNR}, \gls{CNR}, and the energy responses of various materials in the scanner.

\subsection{MUSIC2D}

Tomographic 2D images are easier accessible nowadays and due to the previously-mentioned body of literature on luggage- and cargo inspection, even dual-energy data can be obtained upon request from the application domain community. We decided to evaluate and publish  our spectral data as novel contribution because (i) most open data available for segmentation originate from the medical domain or cargo inspection, where segmentation can be steered by strong shape priors (which is in contrast to actual material identification), and (ii) spectral tomographic data beyond dual energy, covering larger parts of the x-ray spectrum, are rare.

\begin{figure*}[!htbp]
\centering
\subfloat{\includegraphics[width=0.98\textwidth]{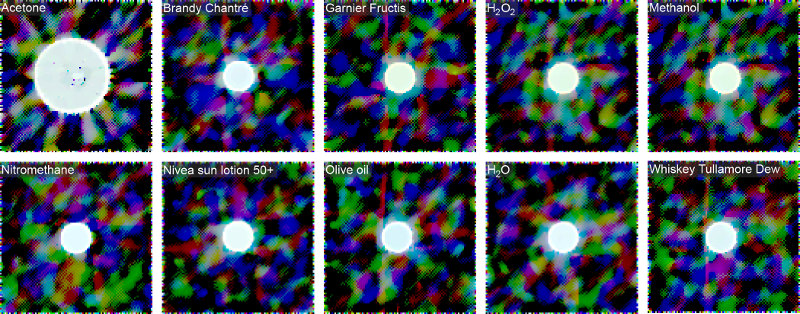}%
\label{fig:music2D:collageAttenuation:refMats}}
\hfil
\subfloat{\includegraphics[width=0.98\textwidth]{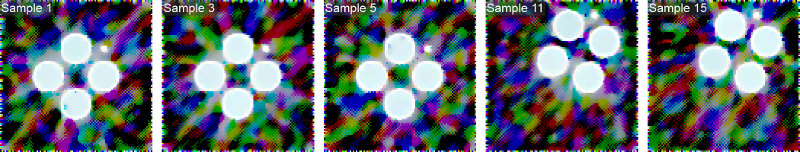}%
\label{fig:music2D:collageAttenuation:4samples}}
\caption{\label{fig:music2D:collageAttenuation} Collection of multi-spectral CT illustrations of MUSIC2D dataset, where each colour channel depicts a specific x-ray energy channel. Note the considerable noise in each energy channel that poses intrinsic challenges to the material identification.}
\end{figure*}

The \textit{MUSIC2D} dataset consists of 32 spectral images in total, of which there are 11 single-object material reference images as material database and 21 multi-object realistic scans to evaluate attenuation interference between the different materials. Fig. \ref{fig:music2D:collageAttenuation} shows a collection of material reference scans and multi-object scans, composed as follows:

\begin{equation}
[R \ G \ B] => [40.52 \ 92.60 \ 120.72] keV
\end{equation}

\begin{figure}[htbp]
\begin{center}
  \centering
  \includegraphics[width=0.98\linewidth]{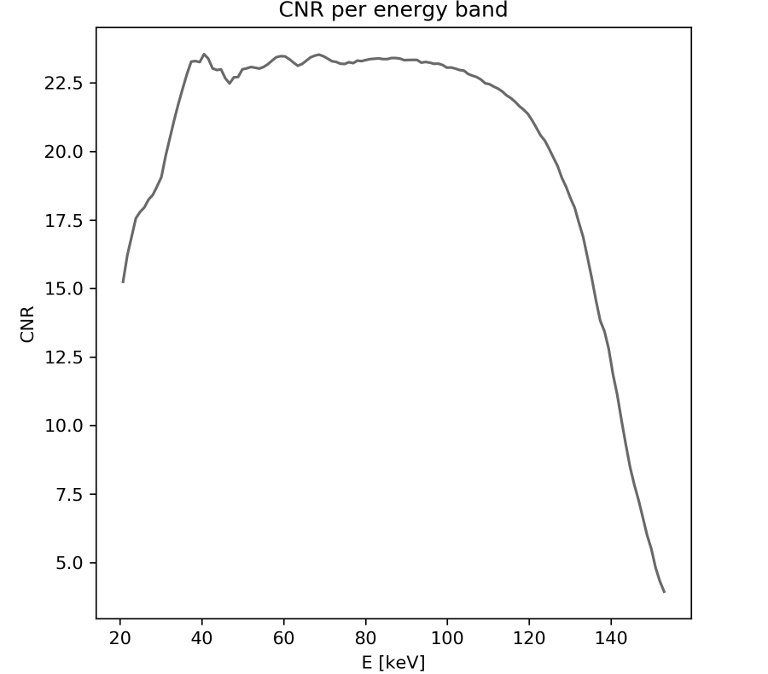}
  \caption{\label{fig:music2D:CNR_refMat} The CNR for the various testing materials (shown as a function of x-ray energy spectrum) shows low signal-to-noise contrast in the spectral regions around 20-35 and 125-160 keV. Conversely, the region close around 30 keV is essential for most fluid identifications.}
\end{center}
\end{figure}

\begin{figure}[htbp]
\begin{center}
  \centering
  \includegraphics[width=0.98\linewidth]{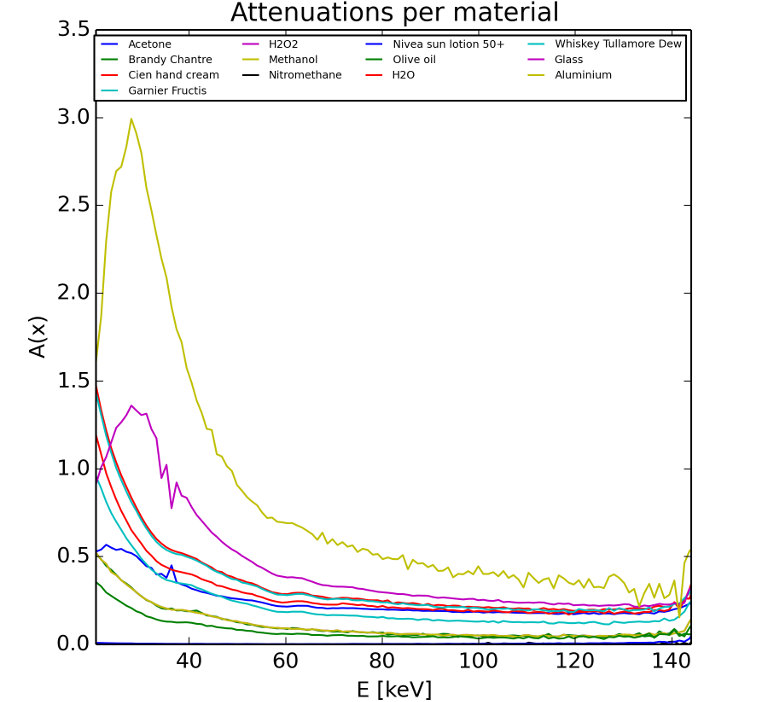}
  \caption{\label{fig:music2D:LAC} The LAC response curves for all 11 reference material scans of MUSIC2D. Notice the photon starvation below 30keV and that the boundary effects from the correction algorithm results in deviations from the physical norm in LACs.}
\end{center}
\end{figure}

Segmenting the data according to their spectral response is challenging without the use of shape priors. The \gls{SNR} within the detector-noise corrected x-ray projections (expressed via non-background photon count and its standard deviation $SNR = \frac{|P_S|}{\sigma_S}$, see \cite{Verdun2015}) is between 0.92 to 1.85 in metalicity-heavy scans and is at 2.34$\pm$0.05 in metal-unaffected scans, while the \gls{SNR} in the reconstruction (measured as $\frac{\bar{f(x)}}{\sigma(f(x))}$ within the image, as in \cite{Behrendt2009}) is between 3.1 (high metalicity) and 4.9 (low metalicity). The average \gls{CNR} (expressed as $\frac{C_{mat}-C_0}{\sigma_0}$, consult \cite{Jaszczak1981} for symbols and definition) is in the range of 3.1 and 6.5. The challenge becomes more apparent when considering the average \gls{CNR} across all 11 single-object scans depending on energy spectrum (fig. \ref{fig:music2D:CNR_refMat}): The low-energy range of 20 keV to 35 keV and high-energy-range of 125 keV to 160 keV shows excessive noise. One control parameter on the \gls{SNR} is the count statistics during the x-ray acquisition (i.e. the amount of emitted and received photons contributing to a pixel's attenuation coefficient), which can be adjusted in the instrument. Low-energy noise artefacts further originate from saturation truncation in the correction algorithm \cite{Dreier2018}, as well as the sample absorption in the radiographic projections (i.e. objects sample one another in the projections, where the object further back receives less photons to detect). High-energy noise is due to generally decreasing count statistics in this part of the spectrum. The typical \gls{LAC} profiles for some tested materials in this article are presented in fig. \ref{fig:music2D:LAC}.

\subsection{MUSIC3D}

The major novelty of this article is the treatment of 3D spectral data (i.e. the \textit{MUSIC3D} dataset). For medical imaging, volumetric phantoms of single-energy CT are openly available whereas multi-energy datasets are not. Additionally, the acquisition of \gls{MECT} for medical diagnosis and treatment is uncommon due to the considerably-increased x-ray exposure of potential patients, and actual patient data is rarely being made public. \gls{MECT} scans are more common in cargo- and luggage assessment, where x-ray exposure of organisms is less of a concern, but openly available datasets are missing while they are in high demand for advanced image analysis. This is the scientific gap filled by MUSIC3D.

\begin{figure}[htbp]
\centering
\subfloat{\includegraphics[width=0.245\linewidth,height=3.6cm,keepaspectratio]{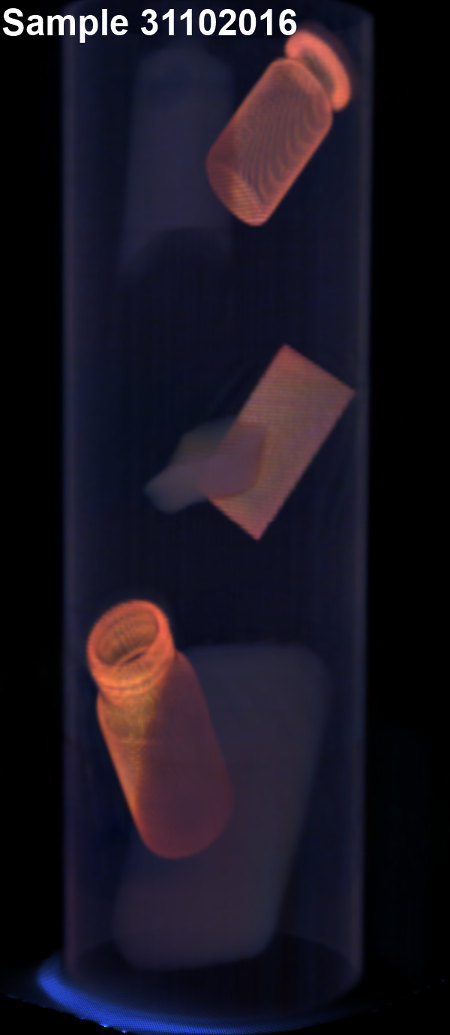}%
\label{fig:music3D:collageAttenuation:Sample_31102016}}
\subfloat{\includegraphics[width=0.245\linewidth,height=3.6cm,keepaspectratio]{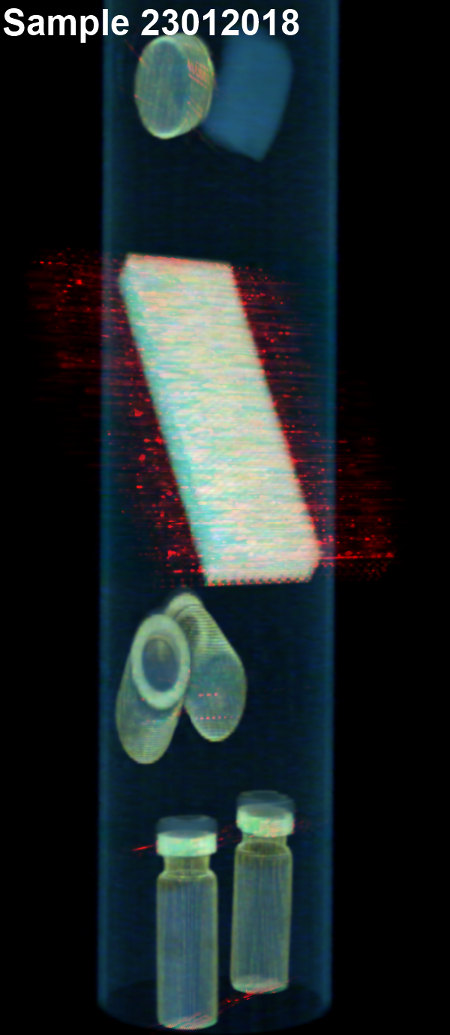}%
\label{fig:music3D:collageAttenuation:Sample_23012018}}
\subfloat{\includegraphics[width=0.245\linewidth,height=3.6cm,keepaspectratio]{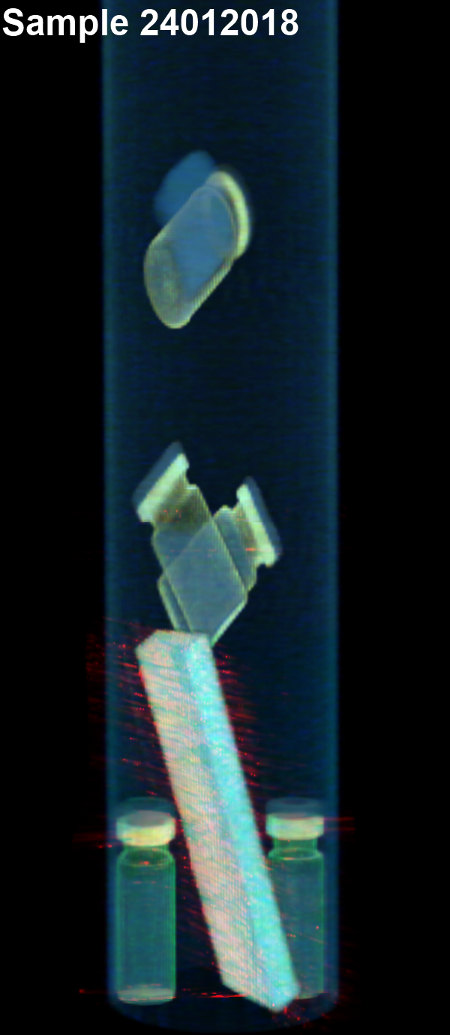}%
\label{fig:music3D:collageAttenuation:Sample_24012018}}
\hfil
\subfloat{\includegraphics[width=0.245\linewidth,height=3.55cm,keepaspectratio]{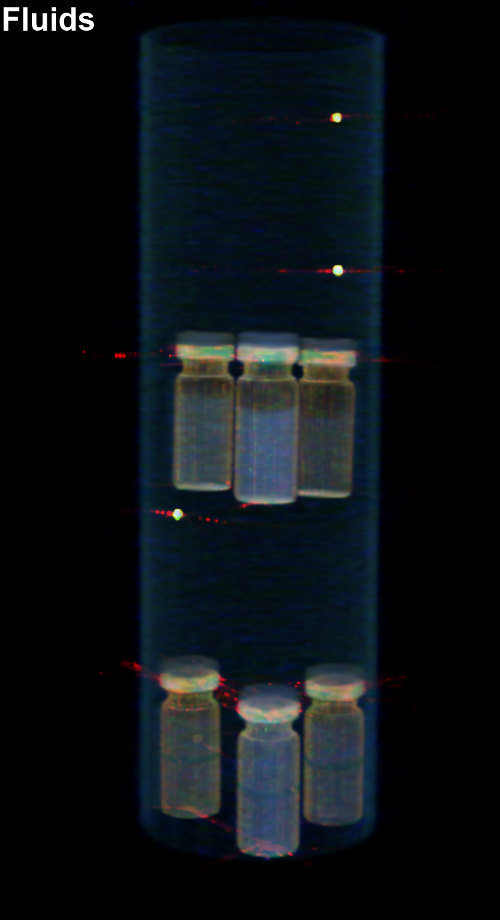}%
\label{fig:music3D:collageAttenuation:Sample_06062018_Fluids}}
\subfloat{\includegraphics[width=0.245\linewidth,height=3.55cm,keepaspectratio]{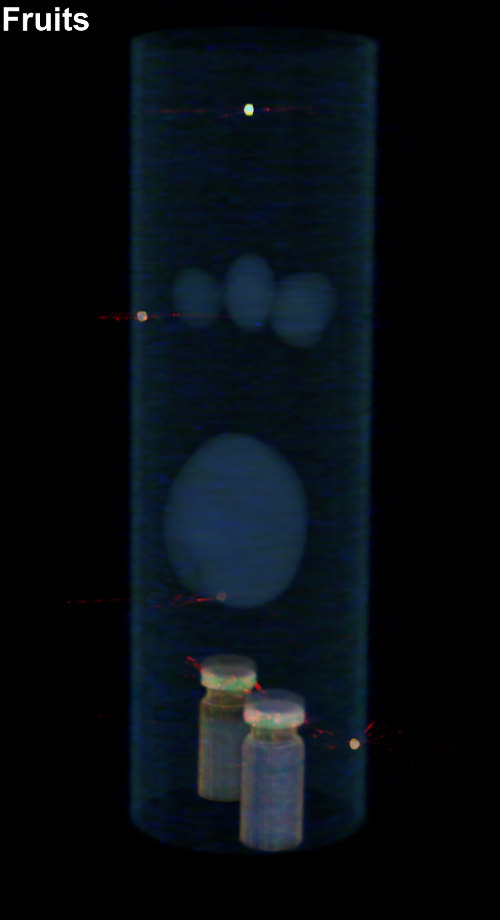}%
\label{fig:music3D:collageAttenuation:Sample_06062018_Fruits}}
\subfloat{\includegraphics[width=0.245\linewidth,height=3.55cm,keepaspectratio]{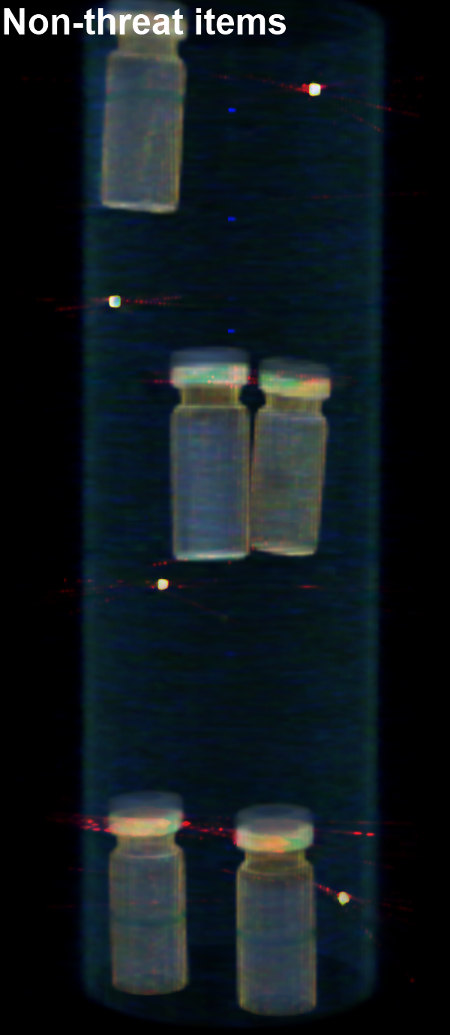}%
\label{fig:music3D:collageAttenuation:Sample_06062018_NonThreat}}
\subfloat{\includegraphics[width=0.245\linewidth,height=3.55cm,keepaspectratio]{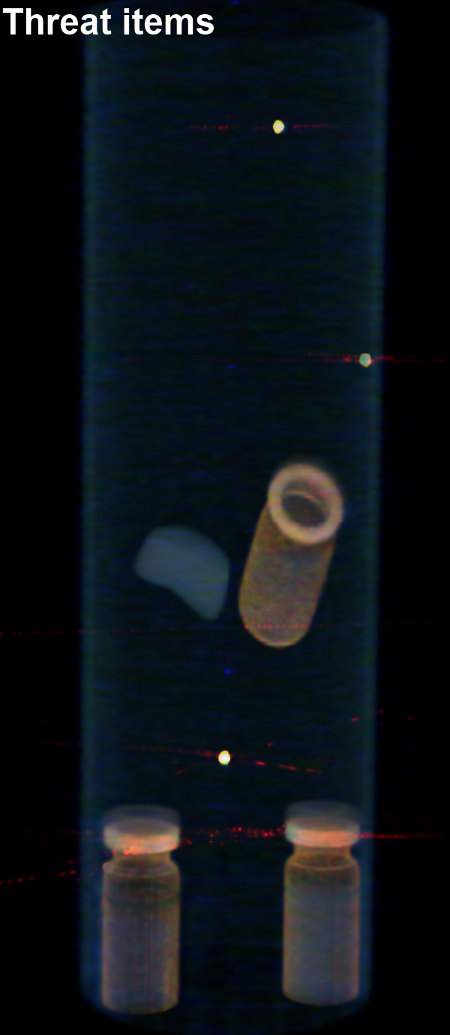}%
\label{fig:music3D:collageAttenuation:Sample_06062018_Threat}}
\caption{Collection of multi-spectral CT illustrations of MUSIC3D dataset, where each colour channel depicts a specific x-ray energy channel.}
\label{fig:music3D:collageAttenuation}
\end{figure}

The MUSIC3D dataset consists of 7 spectral samples in total, all including multiple objects in realistic settings. Two scans (i.e. 'Sample 23012018' and 'Sample 24012018') pose increasing challenges for image segmentation due to an aluminium bar that causes considerable metal artefacts. Fig. \ref{fig:music3D:collageAttenuation} shows the \gls{DVR}-generated images for each scan with the energy channel mapping $[R, G, B] => [40.52, 61.35, 99.89]$ keV.

\section{Algorithms \& Methods}\label{sec:algorithms}

For the segmentation of spectral images and volumes and with respect to the introduced target domain, we quickly summarize key requirements on our target methodology:

\begin{itemize}
\item minimize inner-segment heterogeneity while maximising inter-segment heterogeneity
\item make explicit use of the extra, spectral data dimension
\item not require user input for the  approximate number or locations of target segments
\item not require \textit{a priori} knowledge about the number of objects or materials within the data (i.e. fully unsupervised segmentation)
\item not rely on rigid boundaries or prominent shapes for the segmentation
\item reduce bias by not expecting specific materials to be present in the data
\end{itemize}

The boundary condition on this spectral segmentation are very strict and thus leave few existing algorithm to be applicable to the task, which are further discussed.

%
%
%

\subsection{Spectral Fast Adaptive Mean Shift}\label{sec:algorithms:FAMS}

The mean shift algorithm \cite{Comaniciu2002} is an established tool for multivariate data analysis and has been applied to multi-spectral image analysis in many application domains \cite{Gilani2009,Jordan2013,Banerjee2014}. The algorithm aims at estimating the data density in $\mathds{R}^d$ using a kernel function $K(\mathrm{x})$ (see eq. \ref{eq::fams::kernel}).

\begin{equation}
\label{eq::fams::kernel}
K(\mathrm{x_i}) = c_{k,d}k(||\mathrm{x_i}||^2),
\end{equation}

, with $\mathrm{x_i} \in \mathds{R}^d \ \forall i \in \{1;n\}$ being the data sample, $c_{k,d}$ being a kernel normalisation constant in $\mathds{R}^d$ and $k(x)$ being the kernel profile. With this density estimation, a mean shift vector is constructed, as in eq. \ref{eq::fams::vector_nonadaptive}.

\begin{equation}
\label{eq::fams::vector_nonadaptive}
\mathrm{m}_{h,G}(\mathrm{x}) = \frac{\sum_{i=1}^{n} \mathrm{x_i} g(||\frac{\mathrm{x-x_i}}{h}||^2)}{\sum_{i=1}^{n} g(||\frac{\mathrm{x-x_i}}{h}||^2)}-\mathrm{x_i},
\end{equation}

, with $g(x)$ being the kernel function and $\mathrm{x}$ being the kernel center. Within the resulting data mapping function in $\mathds{R}^d$, the algorithm locates extremal points whose density derivative equals zero (eq. \ref{eq::fams::mode}). These extremal points are called \textit{modes} and are the data cluster centroids in kernel space. In an image segmentation scenario, the modes are segment centres in kernel space and the (pruned) number of modes equals the number of segments.

\begin{equation}
\label{eq::fams::mode}
\nabla \hat{f}_{h,K}(\mathrm{y}_c)=0
\end{equation}

A drawback of the initial equation is the use of a constant kernel size $h$ (eq.~\ref{eq::fams::vector_nonadaptive}), which assumes an isotropic value distribution in $\mathds{R}^d$. From our initial experiments and the analysis of statistical distribution of x-ray intensities, we see that the isotropic value sampling does not apply to \gls{MECT} data. Thus, we apply the \gls{FAMS} extension \cite{Georgescu2003} of \gls{MS} to our data, where the mean shift vector follows the formulation in eq. \ref{eq::fams::vector_adaptive}.

\begin{equation}
\label{eq::fams::vector_adaptive}
\mathrm{m}_G(\mathrm{x}) = \frac{\sum_{i=1}^{n} \frac{1}{h_{i}^{d+2}}  \mathrm{x_i} g(||\frac{\mathrm{x-x_i}}{h_i}||^2)}{\sum_{i=1}^{n} \frac{1}{h_{i}^{d+2}} g(||\frac{\mathrm{x-x_i}}{h_i}||^2)}-\mathrm{x_i}
\end{equation}

In this formulation, the mean shift vector does not depend on a globally-estimated, isotropic bandwidth $h$, but adapts to local anisotropic density variations with an individual density bandwidth per datum $h_i$. Furthermore, Jordan and Angelopoulou proposed using the spectral gradient as input to the mean shift (as opposed to the actual absorption intensities in their spectral photographs) to obtain more coherent segmentations \cite{Jordan2013}. After initial experimental comparisons between mean shifting x-ray intensities or spectral gradients thereof, we decided using the spectral gradient adaptation.

We implemented the algorithm using the reference implementation of \cite{Georgescu2003} in C and wrapped its functionality into Python to provide an easy interface to the actual data processing. The spectral gradient computation is performed within Numpy, as is the uniform data normalisation and quantization, which is necessary as the \gls{FAMS} procedure was specifically designed for integer numerics. 

\subsection{Spectral Graph Cut}\label{sec:algorithms:graphCut}

For the unconditioned graph cut, we use the method formulation by Felzenszwalb and Huttenlocher \cite{Felzenszwalb2004}: Let the image (or volume) be represented by a undirected graph $G=(V,E)$ with each pixel (or voxel) being a vertex $v \in V$ of the graph. Each vertex contains a property vector $f(v) \in \mathds{R}^d$ representing the x-ray intensities or \glspl{LAC}. All vertices are connected to their 1-ring neighbourhood by edges $e(v_i,v_j) \in E$ where the edge length is given by weights $w(e) \in \mathds{R}$ with $w(e) = 0 \ \forall e(v_i, v_j), i=j$. Otherwise, we define the edge weights as the mean vector derivative of two vertices, so that

\begin{equation}
\forall e \in E, w(e) = \Delta f(v_i, v_j) = |f(v_j)-f(v_i)|
\end{equation}

The implications of different 1-ring neighbourhood definitions is discussed with the experiment results.

After the definition of the data terms, our segmentation proceeds as originally described in \cite{Felzenszwalb2004}: each vertex is initialised with a unique segment number. At each iteration, the internal difference per segment $C$ is computed as the maximum weight of its \gls{MST} $MST(C,E)$ (eq. \ref{eq::gcut:internal}). The inter-segment difference is computed as the minimum edge weight connecting two segments (eq. \ref{eq::gcut:difference}). An inter-segment boundary is indicated by the delta-function $D(C_i,C_j)$ (eq. \ref{eq::gcut:delta}). Segments not separated by a boundary are merged. The algorithm terminates when all segments are separated by boundaries. This formulation strictly follows the first criterion laid out for our segmentation goals. Note that these boundaries are limits of the spectral signature and not expected shape boundaries.

\begin{equation}
\label{eq::gcut:internal}
Int(C) = \text{max} \ w(e) \ \forall e \in MST(C,E)
\end{equation}
\begin{equation}
\label{eq::gcut:difference}
Dif(C_i,C_j) = \text{min} \ w(e) \ \forall e \in E, e = e(v_i \in C_i, v_j \in C_j)
\end{equation}
\begin{equation}
\label{eq::gcut:delta}
D(C_i,C_j) = \begin{cases}
true  & \quad \text{if } Dif(C_i,C_j) > MInt(C_i, C_j)\\
false & \quad \text{otherwise}
\end{cases}
\end{equation}

\begin{figure}[htbp]
\begin{center}
	\subfloat[7-neighbourhood]{\includegraphics[width=0.325\linewidth,height=3.5cm,keepaspectratio]{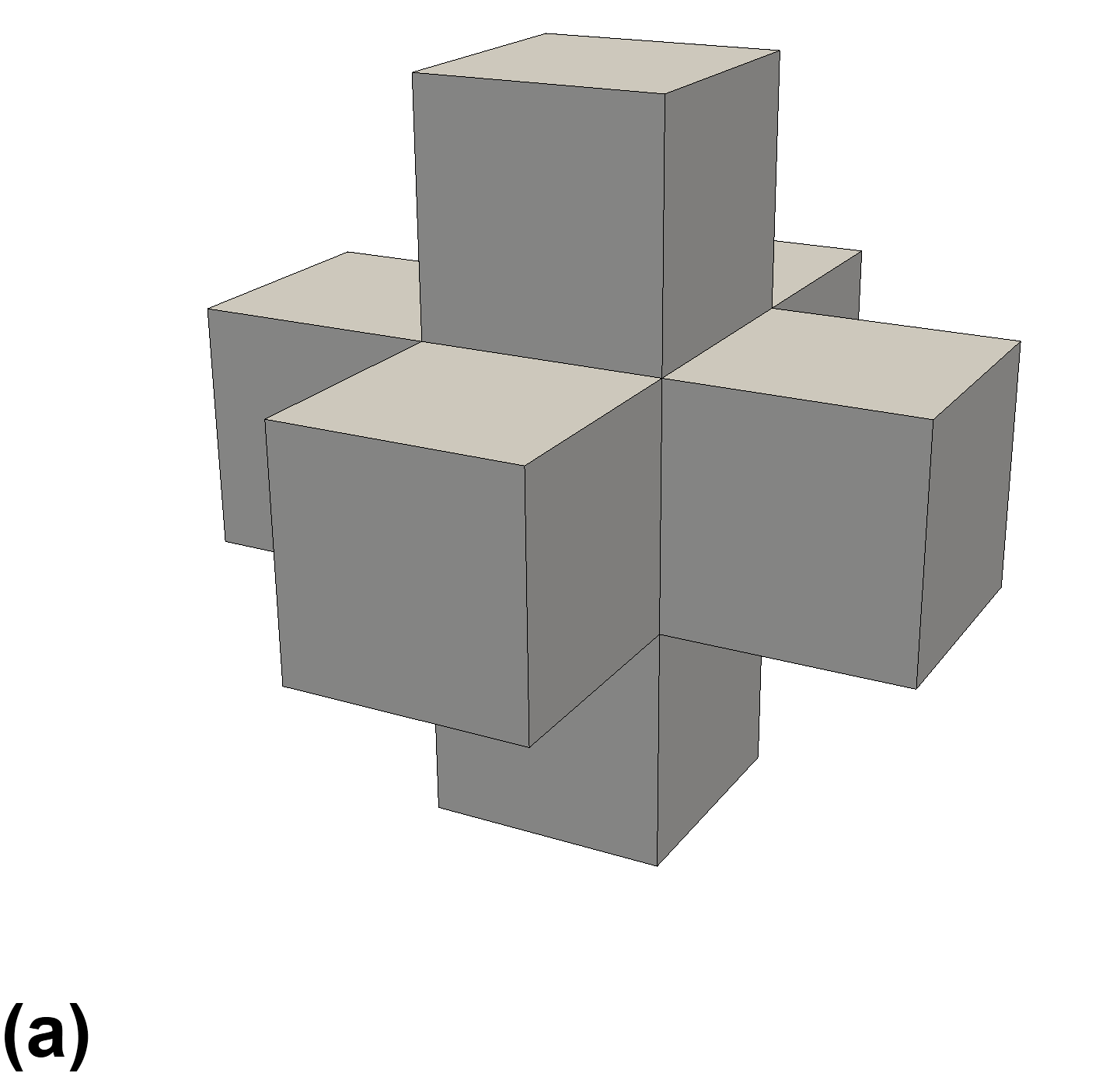} \label{fig:music3D:nbr:seven}}
	\subfloat[27-neighbourhood]{\includegraphics[width=0.325\linewidth,height=3.5cm,keepaspectratio]{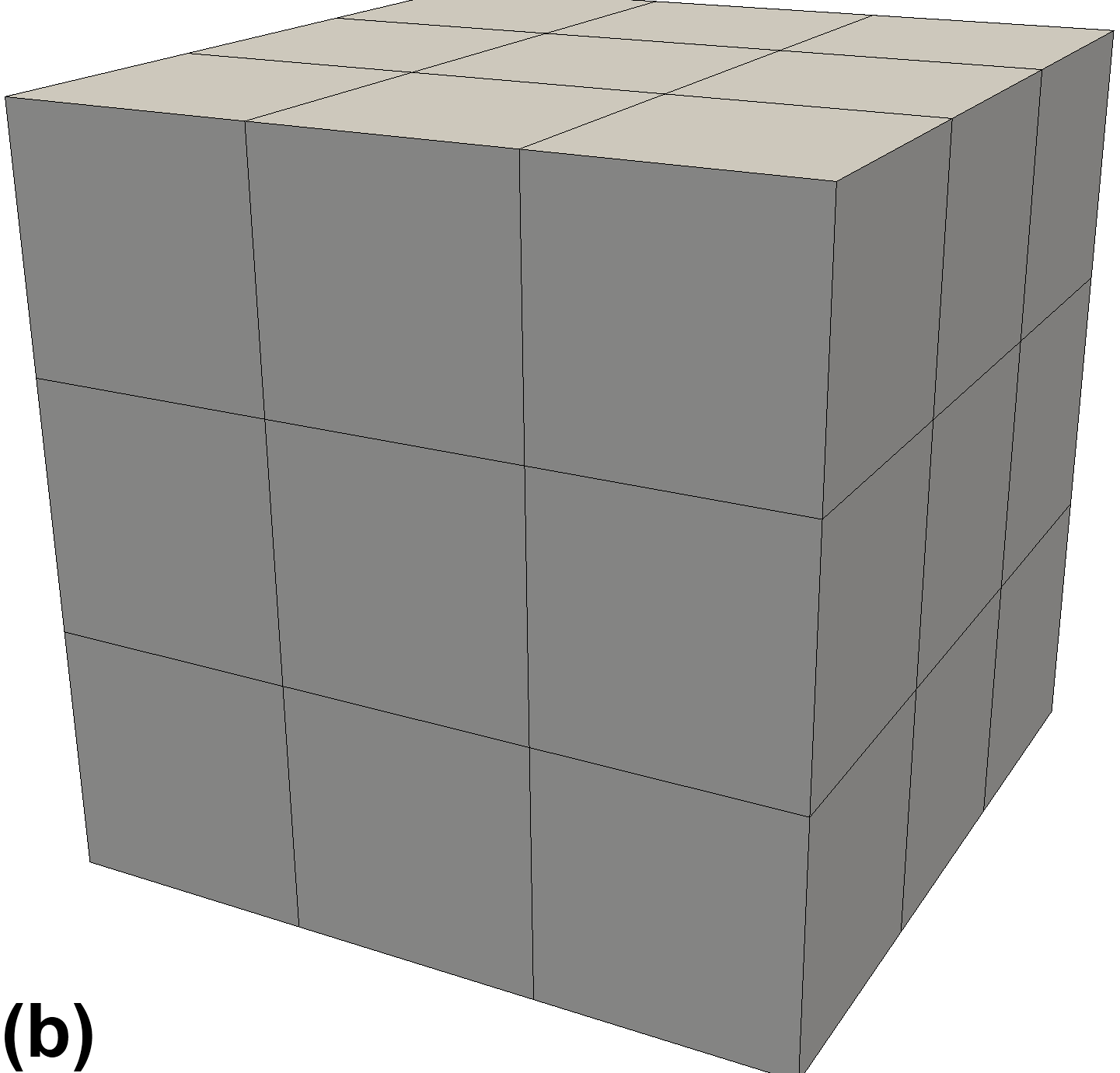} \label{fig:music3D:nbr:twentyseven}}
	\subfloat[weighted 27-neighbourhood]{\includegraphics[width=0.325\linewidth,height=3.5cm,keepaspectratio]{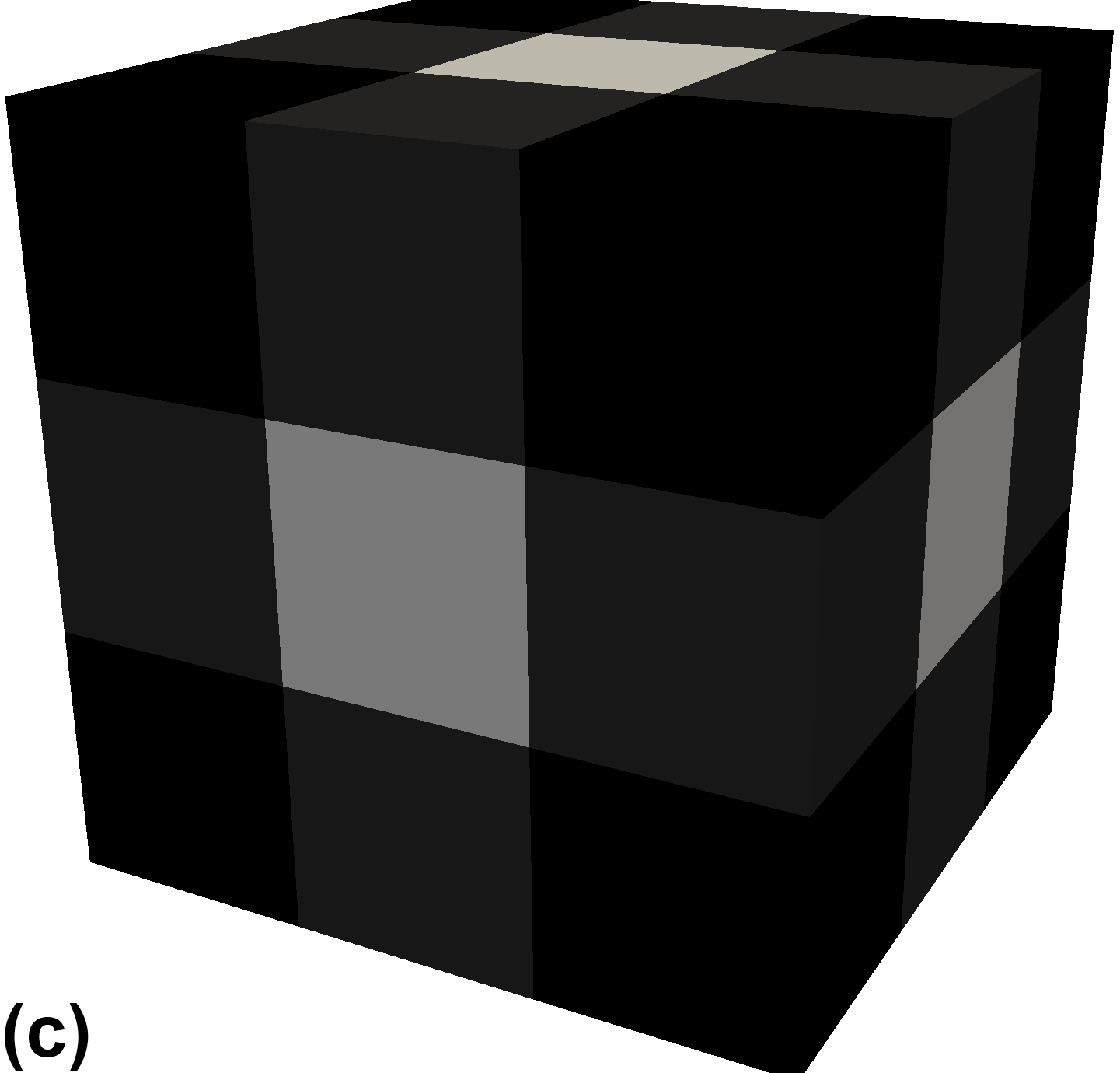} \label{fig:music3D:nbr:twentyseven_weight}}
  \caption{\label{fig:music3D:nbr} Neighbourhood definition illustration. For the weighted neighbourhood, the block intensity maps to the contribution of each adjacent cell.}
\end{center}
\end{figure}

The chosen neighbourhood definition applied to the edge connectivity in graph $G(E,V)$ impacts the segmentation results because the weights of the edges $E$ are composed of neighbouring absolute voxel differences. The original implementation uses a box filter kernel (27-neighbourhood in 3D grids) as connectivity representation (fig. \ref{fig:music3D:nbr}b). This leads to leakage between segments as diagonal boundaries are not respected for containing a segment. We experimented with the constrained 7-neighbourhood definition for vertex adjacency (fig. \ref{fig:music3D:nbr}a), which prevents leakage explicitly but results in a larger number of isolated segments, as well as a bell-curve weighted box filter (i.e. weighted 27-neighbourhood; fig. \ref{fig:music3D:nbr}c), which represents an adaptable trade-off between segment leakage and segment isolation.

\subsection{Adaptive, anisotropic binning}\label{sec:algorithms:adaptiveBinning}

The initial segmentation results, especially in 3D spectral domains, have shown unsatisfying results for subsequent material identification, even when omitting the high-noise bottom- and top end of the acquired x-ray spectrum (see \gls{CNR} in fig. \ref{fig:music2D:CNR_refMat}). A detailed analysis of the attenuation profiles in the 3D spectral data revealed that the material information most distinct for fluids are located in narrow bands of the x-ray spectrum, located between 38 keV and 70 keV (see  fig. \ref{fig:music2D:LAC}). The channels in focus for material distinction are in a low-to-mid energy range where solid materials as well as liquids are significantly attenuating the penetrating x-rays (i.e. more than 30\% \cite{Vopalensky2017}). Separating different solid materials (e.g. plastics, metals) can be done in higher energy channels too, while liquids are attenuating the high-energy x-rays to little to be visible in the upper spectral range. If applying uniform (i.e. \textit{isotropic}) sampling of the energy bins, these low-energy high-information channels are accumulated into one energy bin, leading to a loss of information.

We correct the information loss by introducing an adaptive, anisotropic binning scheme that reserves more output bins for the high-variability bands in the x-ray spectrum. The adaptive binning computes the variance $\sigma(f(x,e))$ for each input energy channel separately, as well as the global sum $\sigma(f(x))$ of all energy levels. Then, we define the budget $L(e_{out})$ for each output bin as follows:

\begin{equation}
L(e_{out}) \leq \frac{\sigma(f(x))}{N},
\end{equation}

with $N$ being the number of output bins. According to this budget, energy channels are accumulated while the distinct energy information are preserved relative to overall data variability.



\section{Results}
\label{sec:results}

In this section, we present the segmentation results obtained with the methods discussed above. A discussion of the \gls{FAMS} and unsupervised graph cut segmentation with minimal data modification is followed by an in-depth analysis of the impact of adaptive spectral binning. Moreover, we discuss how the deteriorating quality of limited-projection tomographic reconstruction influences the segmentation results on the adaptively-binned spectral datasets.

The two segmentation methods utilised in this study each evaluate the spectral gradient: the graph cut methods does so explicitly via edge weights, while the mean shift does so implicitly by localising extremal points in the feature plane, which is here described by the spectral dimension. As there is too few variation between each individual channel response and because the data need to be trimmed to avoid harmful frequencies (see section \ref{sec:literature:reconstruction}), the full spectrum data are first clipped between 35 keV and 140 keV and then uniformly rebinned into 20 averaged channels.

\subsection{FAMS segmentation}
\label{sec:results:fams}

\begin{figure}[htbp]
\begin{center}
  \centering
  \includegraphics[width=0.9\linewidth]{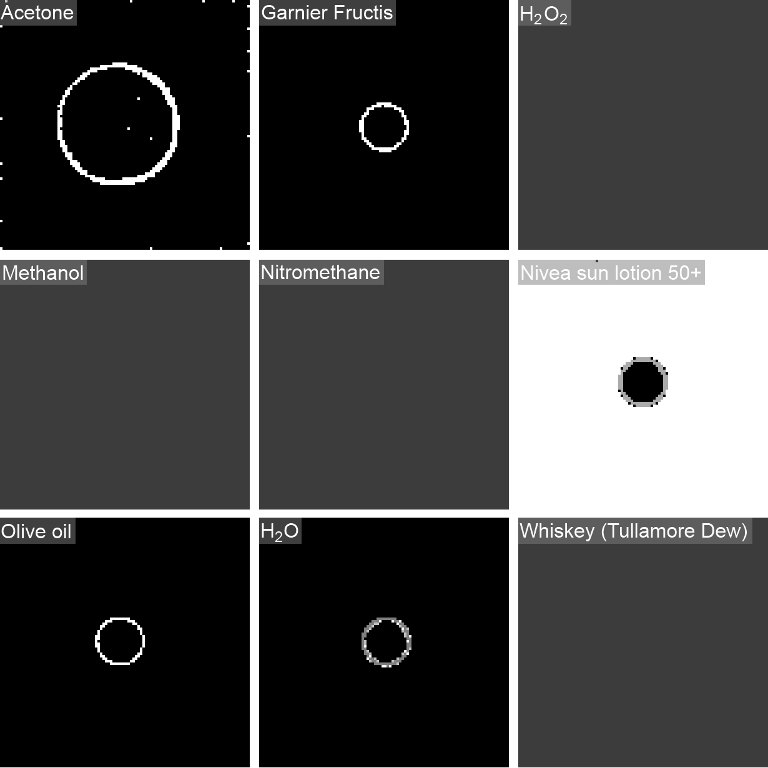}
  \caption{\label{fig:music2D:FAMS:refMats} Collection of the FAMS segmentation results achieved on 9 of the available 11 reference material scans. Completely greyed-out images did not provide any segments.}
\end{center}
\end{figure}

\begin{figure}[htbp]
\begin{center}
  \centering
  \includegraphics[width=0.9\linewidth]{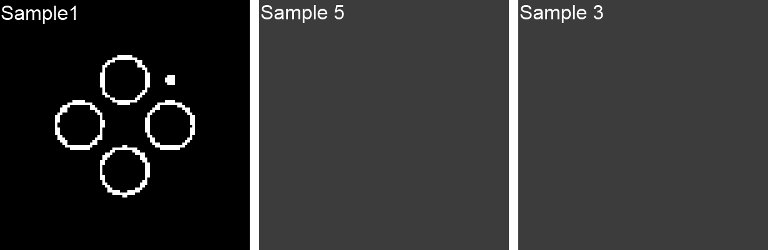}
  \caption{\label{fig:music2D:FAMS:4samples} Collection of the FAMS segmentation results achieved on 3 of the available 20 multi-object scans. Completely greyed-out images did not provide any segments. The composition of each sample is given in app. \ref{app:multi_obj_compose}.}
\end{center}
\end{figure}

Fig. \ref{fig:music2D:FAMS:refMats} shows the segmentation results on 9 of the 11 single-object scans of the MUSIC2D dataset using \gls{FAMS} across the full spectrum (i.e. uniform binning). Figure \ref{fig:music2D:FAMS:4samples} shows the segmentation maps for 3 selected cases of the multi-object scans in the 2D dataset. As seen in the figures, the spectral \gls{FAMS} is prevalently unable to extract larger segment regions, though it is often able to extract the spectral boundary (i.e. boundary formed by high-intesity spectral gradient) between materials. Hence, while the spectral \gls{FAMS} method fails at providing good segmentations for \gls{MECT} (in contrast to previously-documented literature for visible- and infrared spectra \cite{Angelopoulou1999,Jordan2013}), it provides spectral boundary constraints that can be used to extract connected components and then to derive spectrally-separate objects. Other versions of the local density-adaptive mean shift (also listed in \cite{Jordan2013}) have been tested directly on the recorded energy channels, but no mean shift algorithm was able to extract consistent spectral object maps usable for subsequent material identification.

These 2D results apply the standard parameterization of $k=100$ (neighbourhood query size), $K=24$ (observation size) and $L=35$ (detail level) (consult \cite{Georgescu2003} for parameter descriptions).

%

\begin{figure}[htbp]
\centering
\subfloat{\includegraphics[width=0.245\linewidth,height=4.0cm,keepaspectratio]{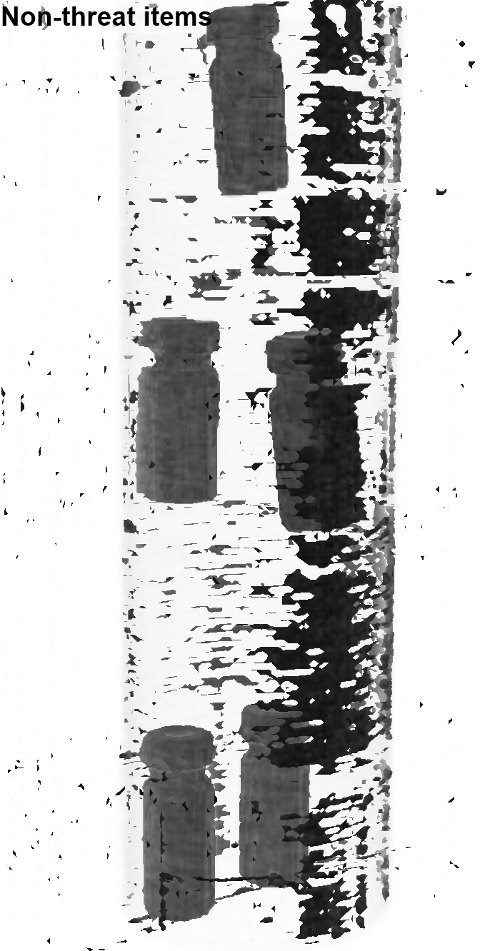}%
\label{fig:music3D:FAMS:Sample_06062018_NonThreat}}
\subfloat{\includegraphics[width=0.245\linewidth,height=4.0cm,keepaspectratio]{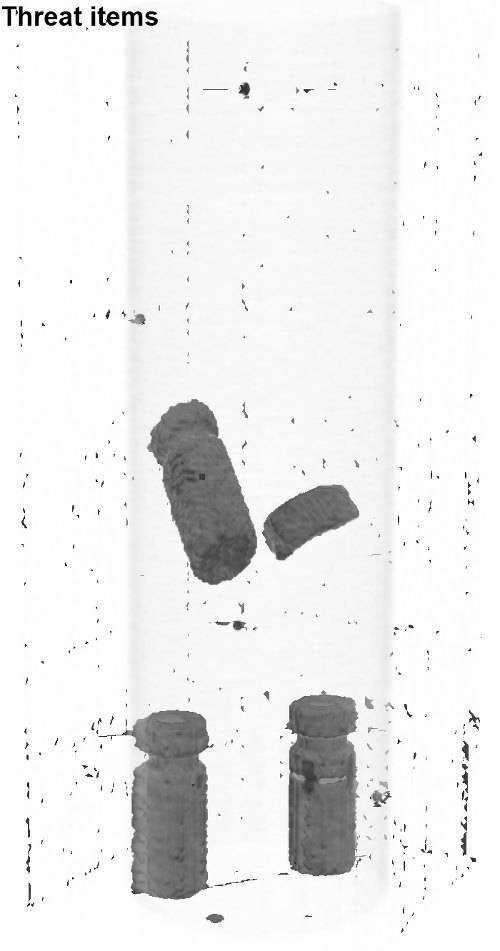}%
\label{fig:music3D:FAMS:Sample_06062018_Threat}}
\subfloat{\includegraphics[width=0.245\linewidth,height=4.0cm,keepaspectratio]{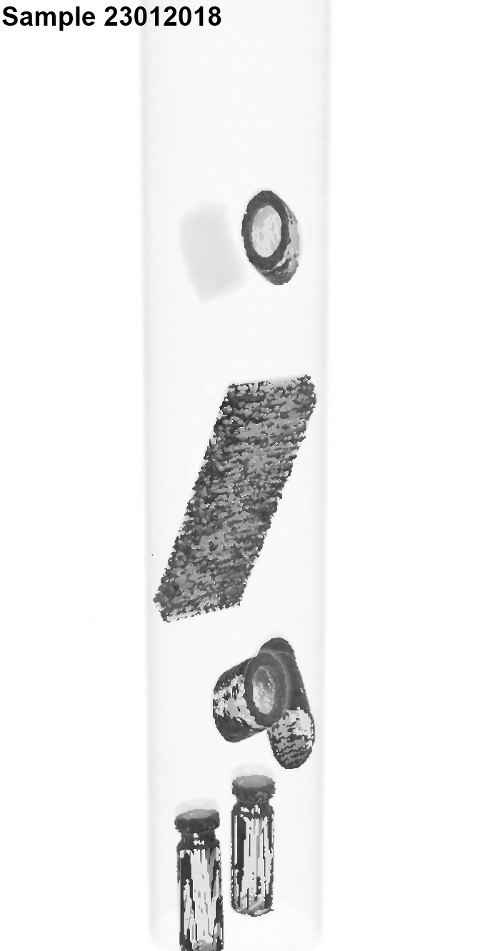}%
\label{fig:music3D:FAMS:Sample_23012018}}
\subfloat{\includegraphics[width=0.245\linewidth,height=4.0cm,keepaspectratio]{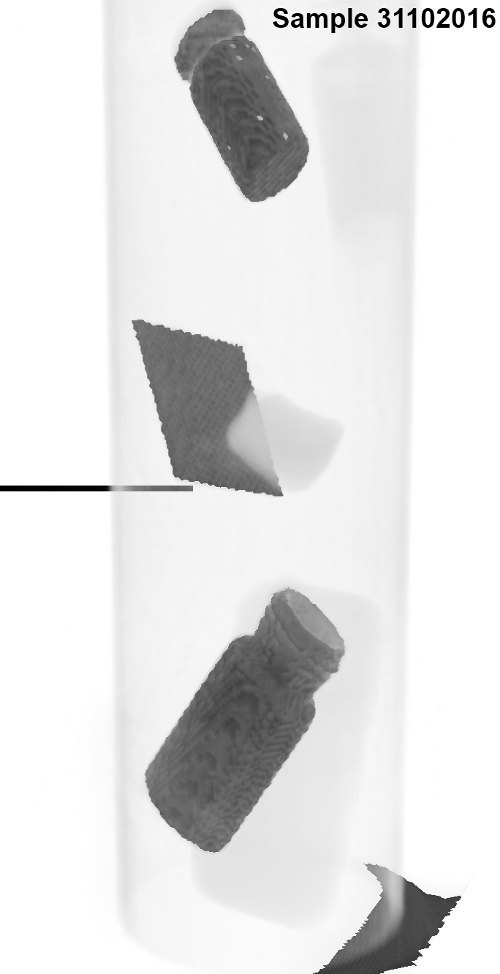}%
\label{fig:music3D:FAMS:Sample_31102016}}
\caption{FAMS segmentation results achieved on the MUSIC3D dataset. Spectral FAMS predominantly extracts the segment boundaries as the target segments (i.e. output is a binary segmentation of boundaries). For some datasets ('Sample24012018','Fluids','Fruits'), their content is indistinguishable from the background attenuation, leading to an empty segment map. The illustration embeds the segments (black colour) in a DVR of energy channel 39 (61.35 keV).}
\label{fig:music3D:FAMS}
\end{figure}

Fig. \ref{fig:music3D:FAMS} provides an overview of the achievable segmentations on the MUSIC3D dataset with the \gls{FAMS} segmentation. As with the 2D spectral data, the segmentation results of 3D data are of insufficient quality for direct material identification. We failed to extract any segments via \gls{FAMS} for 3 samples of the 3D spectral dataset, which is due to the low \gls{SNR}. The method's parameters of $k$, $L$ and $K$ (see their description in \cite{Georgescu2003}) needed to be fine-tuned for each sample to achieve a non-arbitrary segmentation, due to the low \gls{SNR}. When segmenting the 3D datasets, we established the following range of working parameter settings experimentally.

\begin{itemize}
\item k: 100 -- 220 (std.: 100)
\item K: 20 -- 30 (std.: 24)
\item L: 30 -- 40 (std.: 35)
\end{itemize}

\begin{figure}[htbp]
\centering
\subfloat{\includegraphics[width=0.49\linewidth,height=4.1cm,keepaspectratio]{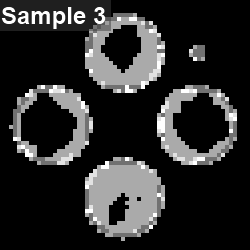}%
\label{fig:music2D:autoFAMS:4samples:sample3}}
\subfloat{\includegraphics[width=0.49\linewidth,height=4.1cm,keepaspectratio]{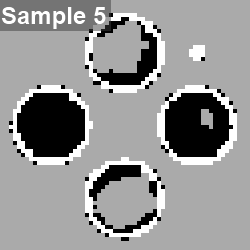}%
\label{fig:music2D:autoFAMS:4samples:sample5}}
\hfil
\subfloat{\includegraphics[width=0.49\linewidth,height=4.1cm,keepaspectratio]{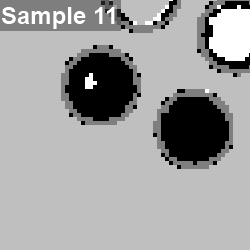}%
\label{fig:music2D:autoFAMS:4samples:sample11}}
\subfloat{\includegraphics[width=0.49\linewidth,height=4.1cm,keepaspectratio]{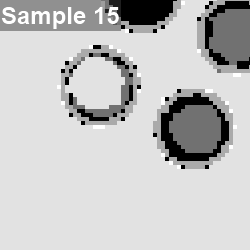}%
\label{fig:music2D:autoFAMS:4samples:sample16}}
\caption{Results of the auto-parameterization on the full-spectrum MUSIC2D FAMS segmentation. The images show results of 'sample3' (k=20,K=15,L=15), 'sample5' (k=150,K=12,L=16), 'sample11' (k=120,K=13,L=17) and 'sample15' (k=100,K=13,L=14), which show improved segmentation of material segments compared to the standard parameterization in fig. \ref{fig:music2D:FAMS:4samples}. All images are zoomed-in version of the full-scale results for better detail visibility.}
\label{fig:music2D:autoFAMS:4samples}
\end{figure}

\begin{figure}[htbp]
\centering
\subfloat{\includegraphics[width=0.325\linewidth]{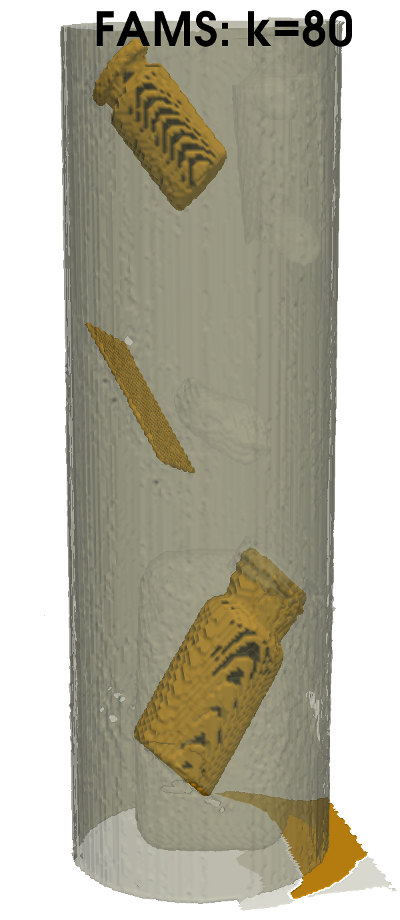}%
\label{fig:music3D:autoFAMS:k080}}
\subfloat{\includegraphics[width=0.325\linewidth]{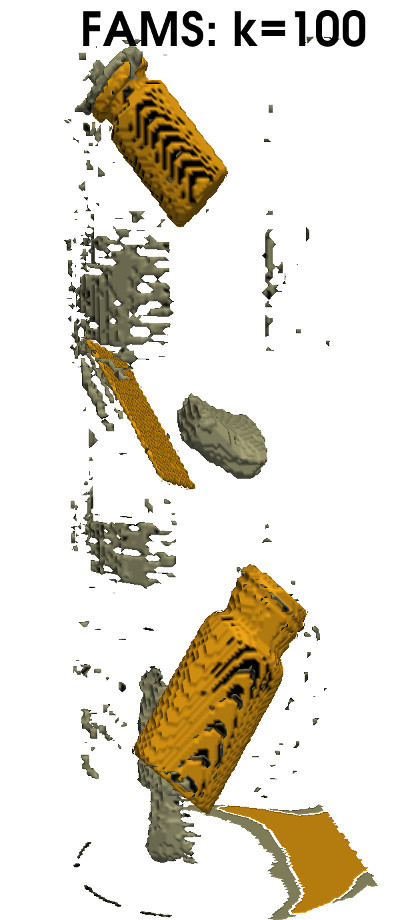}%
\label{fig:music3D:autoFAMS:k100}}
\subfloat{\includegraphics[width=0.325\linewidth]{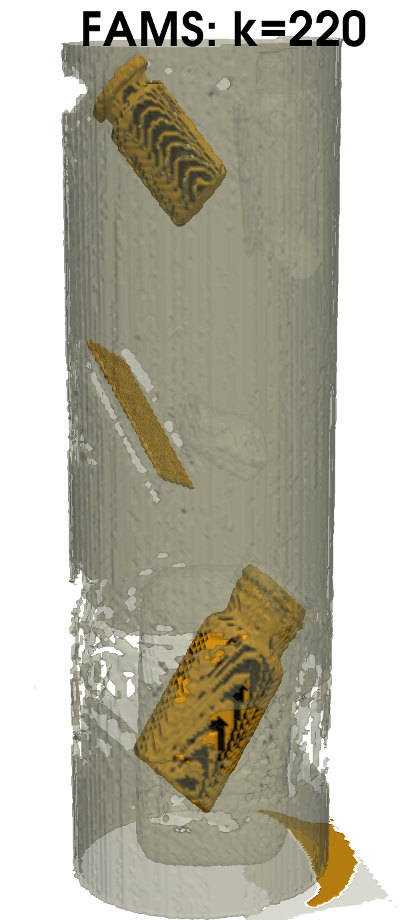}%
\label{fig:music3D:autoFAMS:k220}}
\caption{Result of the auto-parameterization option of FAMS: the result for neighbourhood query sizes of 80, 100 and 220 are shown after optimal observation sizes were determined for 'Sample 31102016'. The algorithm is able to extract object interiors (semi-translucent grey contours) and exteriors (opaque orange contours). The interiors can be used as input for connected component extraction.}
\label{fig:music3D:autoFAMS}
\end{figure}

The drawback of the parameterisation is rooted in the \gls{FAMS} method: it assumes that the data is gaussian distributed and that each value is equiprobable. Due to the underlying decay in attenuation (see fig. \ref{fig:music2D:LAC}) for most non-metallic materials, the input data to \gls{FAMS} from \gls{MECT} is neither gaussian distributed nor is each value occurrence equally probable. This affects the neighbourhood determination via \gls{LSH} \cite{Gionis1999}, upon which \gls{FAMS} is based on, because a neighbour query is more probable to locate neighbours in higher parts of the x-ray spectrum. The speed of the \gls{LSH} query is what drives the observation size parameters $K$ and $L$, which in return results in small observation sizes (and thus: small segments) of \gls{FAMS}. This theory is supported by our experiments: we executed the integrated auto-parameterization provided by Georgescu and Shimshoni \cite{Georgescu2003,Shimshoni2006}, which determines optimal observation sizes depending on the size of the neighbourhood query $k$. The auto-parameterisation yields better results on the full-spectrum MUSIC2D dataset that actually includes multiple material segments (see fig. \ref{fig:music2D:autoFAMS:4samples} for samples with multi-material objects). The results of the auto-parameterization for scan 'Sample 31102016' vary significantly and, in the case of $k=100$, even provide an acceptable objects segmentation (fig. \ref{fig:music3D:autoFAMS}). Despite the potential improvements of the auto-parameterization, the actual segments still need to be extracted via extraction of connected components. Furthermore, because the auto-parameterization relies on execution time measurements, the results of the procedure vary with changing system load during the program's execution.

\subsection{Graph cut segmentation}
\label{sec:results:graphcut}

\begin{figure}[htbp]
\begin{center}
  \centering
  \includegraphics[width=0.98\linewidth]{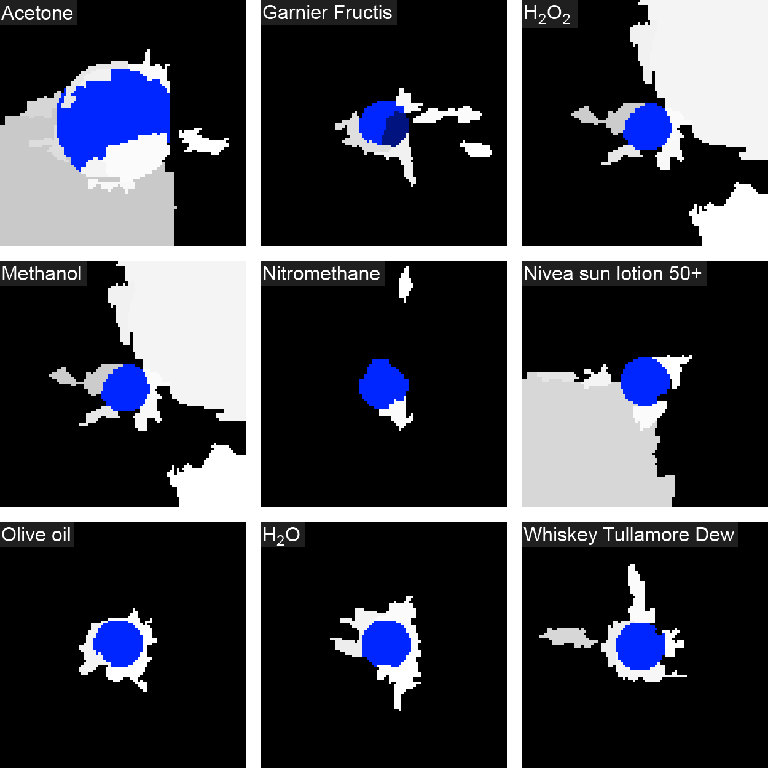}
  \caption{\label{fig:music2D:graphcut:refMats}Renderings of the achievable graph cut segmentation map for 9 of the 11 reference material scans. Segments with equal tones but different brightness are composed of multiple segment indicators.}
\end{center}
\end{figure}

\begin{figure}[htbp]
\centering
\subfloat{\includegraphics[width=0.49\linewidth,height=3.55cm,keepaspectratio]{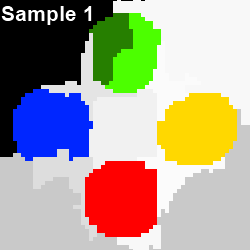}\label{fig:music2D:graphcut:4samples:sample1:seg}}
\subfloat{\includegraphics[width=0.49\linewidth,height=3.55cm,keepaspectratio]{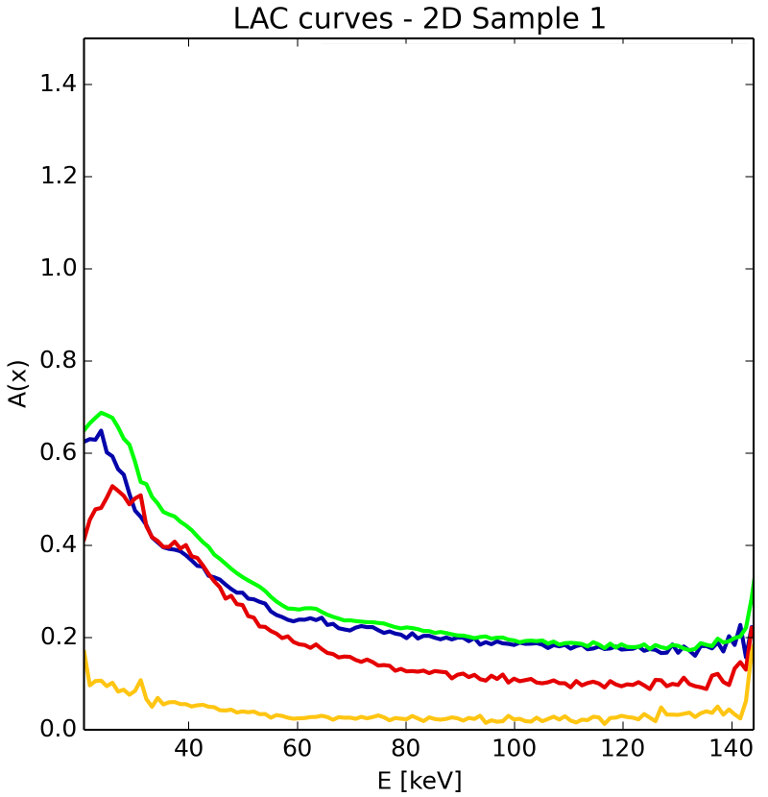}\label{fig:music2D:graphcut:4samples:sample1:lac}}
\hfil
\subfloat{\includegraphics[width=0.49\linewidth,height=3.55cm,keepaspectratio]{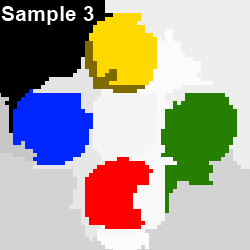}\label{fig:music2D:graphcut:4samples:sample3:seg}}
\subfloat{\includegraphics[width=0.49\linewidth,height=3.55cm,keepaspectratio]{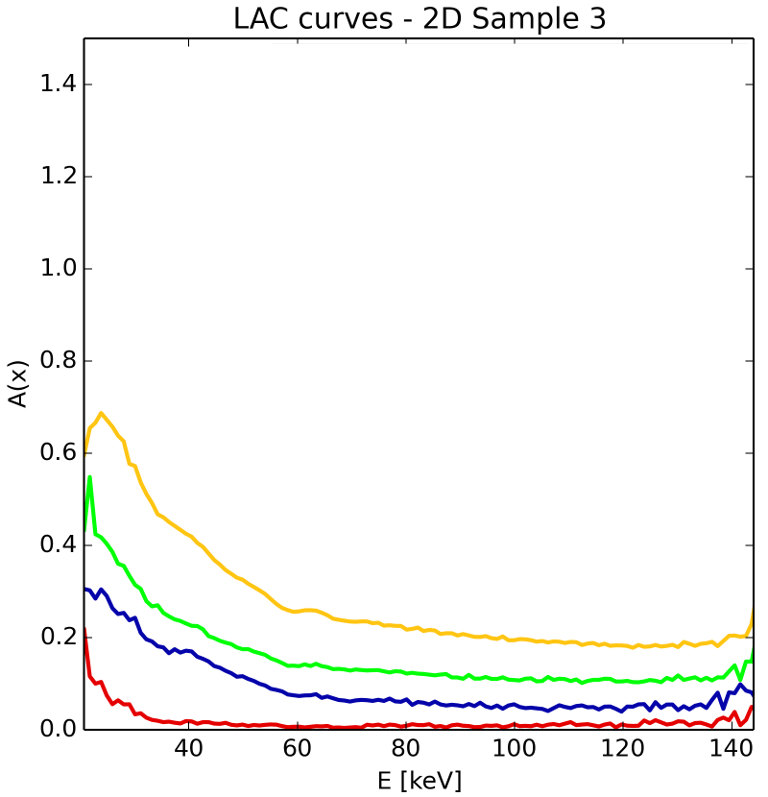}\label{fig:music2D:graphcut:4samples:sample3:lac}}
\hfil
\subfloat{\includegraphics[width=0.49\linewidth,height=3.55cm,keepaspectratio]{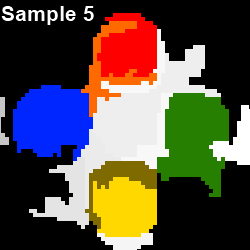}\label{fig:music2D:graphcut:4samples:sample5:seg}}
\subfloat{\includegraphics[width=0.49\linewidth,height=3.55cm,keepaspectratio]{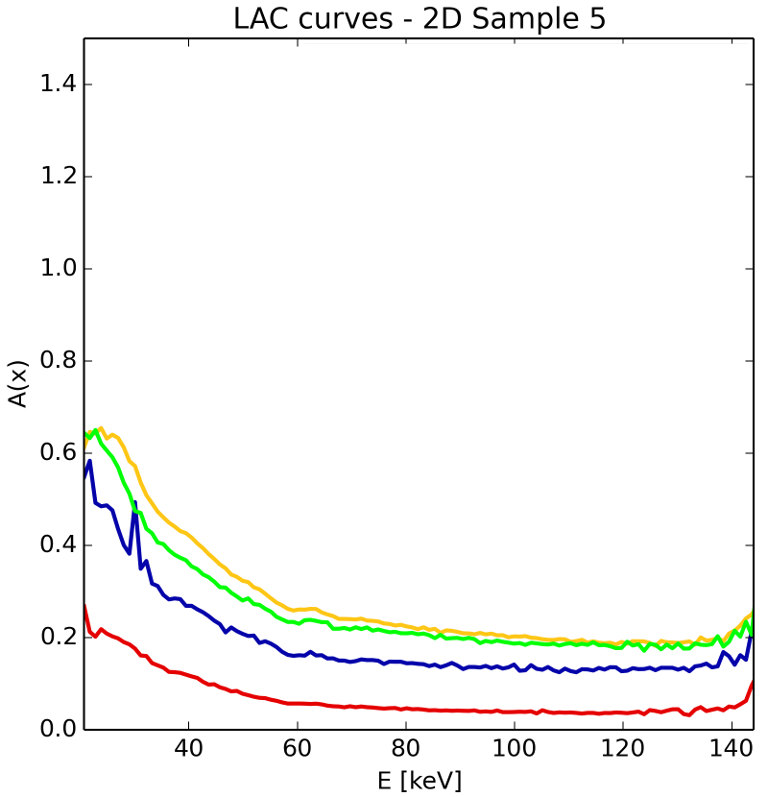}\label{fig:music2D:graphcut:4samples:sample5:lac}}
\caption{Side-by-side comparison of the graph cut segments and LAC responses per material for 3 multi-object scans (for sample composition, see app. \ref{app:multi_obj_compose}). Segments with equal tones but different brightness are composed of multiple segment indicators.}
\label{fig:music2D:graphcut:4samples}
\end{figure}

In comparison to the spectral \gls{FAMS} method, the unsupervised spectral graph cut yields more appropriate segmentation results. Figures \ref{fig:music2D:graphcut:refMats} and \ref{fig:music2D:graphcut:4samples} display the segmentation maps for the 9 single-material reference samples of MUSIC2D and side-by-side comparisons of \glspl{LAC} and graph cut segment maps for 3 multi-object scans in the same 2D dataset. The material boundaries still show considerable uncertainties, but the segments are visually of acceptable quality so to be used for subsequent material identification.

Table \ref{tab:music2D:dice} presents the dice coefficient as quantitative evaluation of segmentation quality for each scan in the MUSIC2D dataset for the graph cut results. The manual reference segmentation for each scan has been obtained manually beforehand using \gls{MITK} \cite{Wolf2005} and \gls{DeVIDE} \cite{Botha2008}.


\begin{table}[h]
\caption{Dice coefficient overview for MUSIC2D dataset as quantitative graph cut segmentation quality evaluation}
\label{tab:music2D:dice}
\centering
\begin{tabular}{|l||r|}
\hline
\textbf{Sub-dataset} & \textbf{dice coeff.}\\
\hline
Acetone            & 0.7023\\
\hline
Brandy Chantr\'{e} & 0.5879\\
\hline
Cien hand cream    & 0.8145\\
\hline
Garnier Fructis    & 0.5760\\
\hline
H$_2$O$_2$         & 0.8290\\
\hline
Methanol           & 0.4573\\
\hline
Nitromethane       & 0.5930\\
\hline
Nivea sun lotion 50+ & 0.5617\\
\hline
Olive oil          & 0.5340\\
\hline
H$_2$O             & 0.8694\\
\hline
Whiskey Tullamore Dew & 0.5844\\
\hline
\hline
Overall            & 0.6463\\
\hline
\end{tabular}
\end{table}

\begin{figure}[htbp]
\centering
\subfloat{\includegraphics[width=0.245\linewidth,height=4.0cm,keepaspectratio]{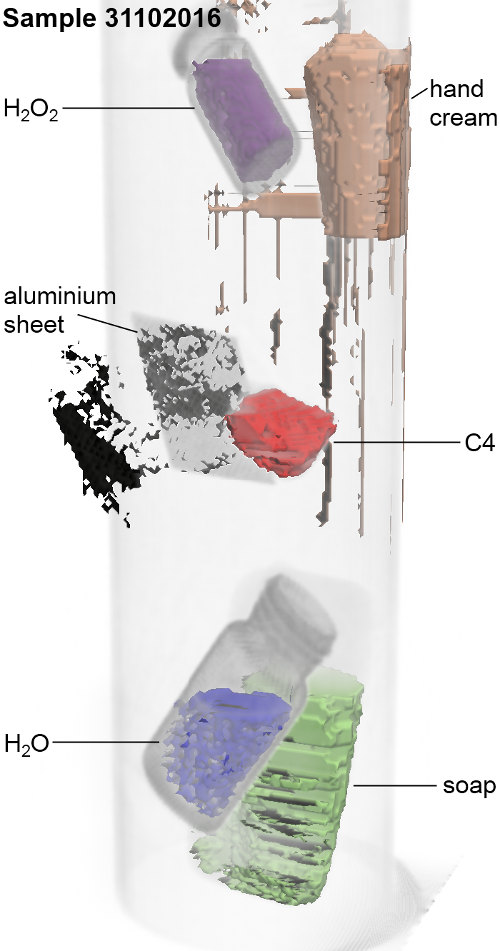}%
\label{fig:music3D:graphCut:Sample_31102016}}
\subfloat{\includegraphics[width=0.245\linewidth,height=4.0cm,keepaspectratio]{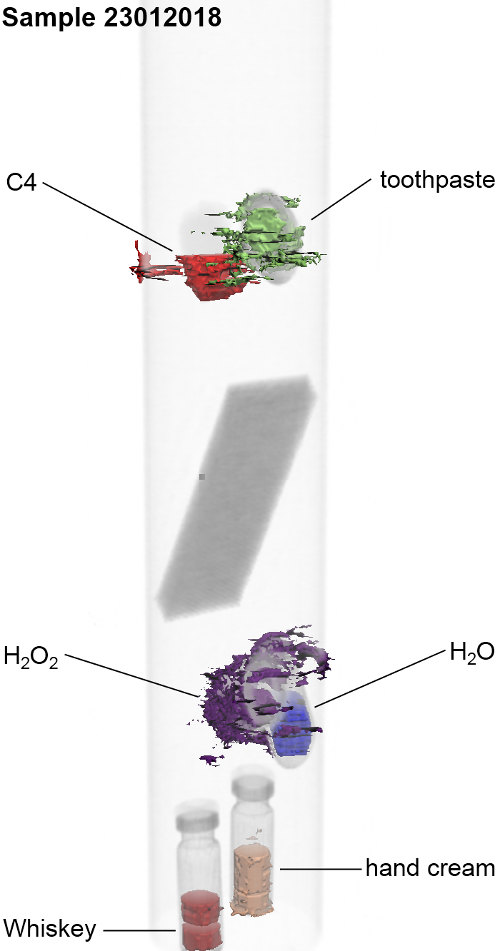}%
\label{fig:music3D:graphCut:Sample_23012018}}
\subfloat{\includegraphics[width=0.245\linewidth,height=4.0cm,keepaspectratio]{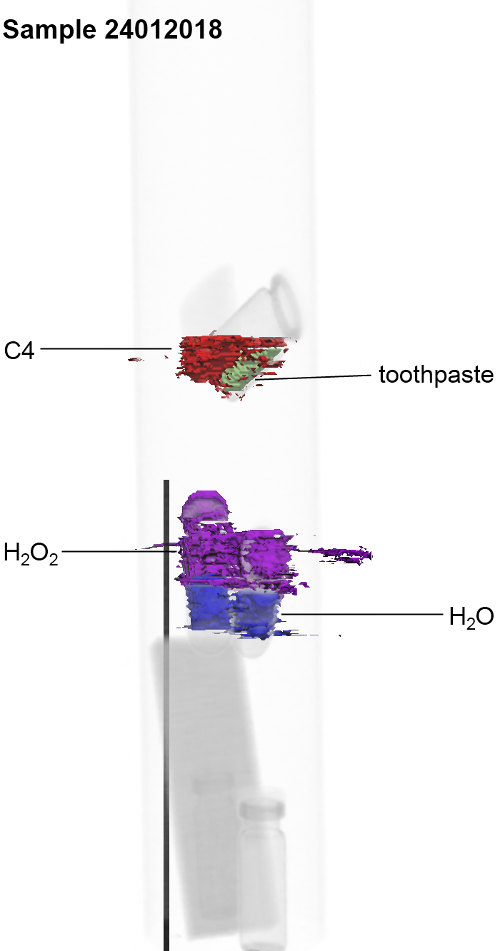}%
\label{fig:music3D:graphCut:Sample_24012018}}
\subfloat{\includegraphics[width=0.245\linewidth,height=4.0cm,keepaspectratio]{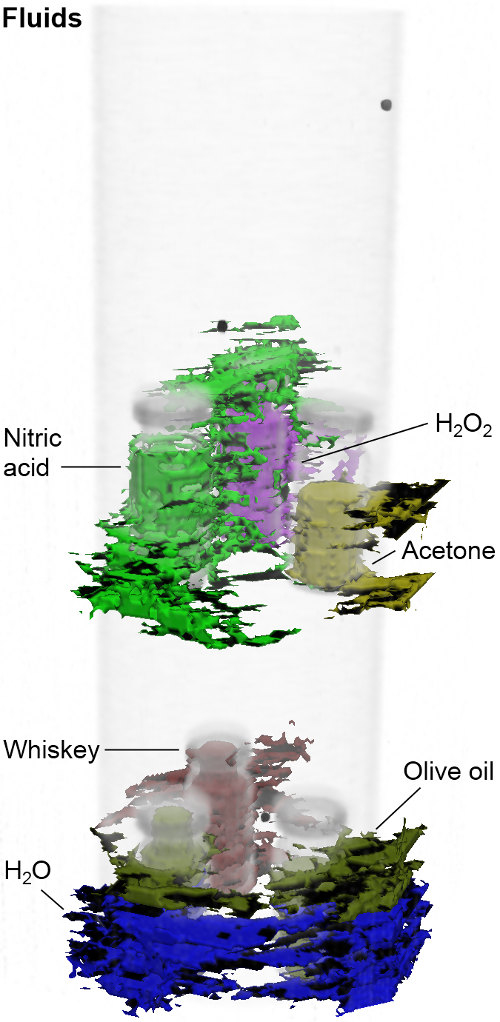}%
\label{fig:music3D:graphCut:Sample_06062018_Fluids}}
\hfil
\subfloat{\includegraphics[width=0.245\linewidth,height=4.0cm,keepaspectratio]{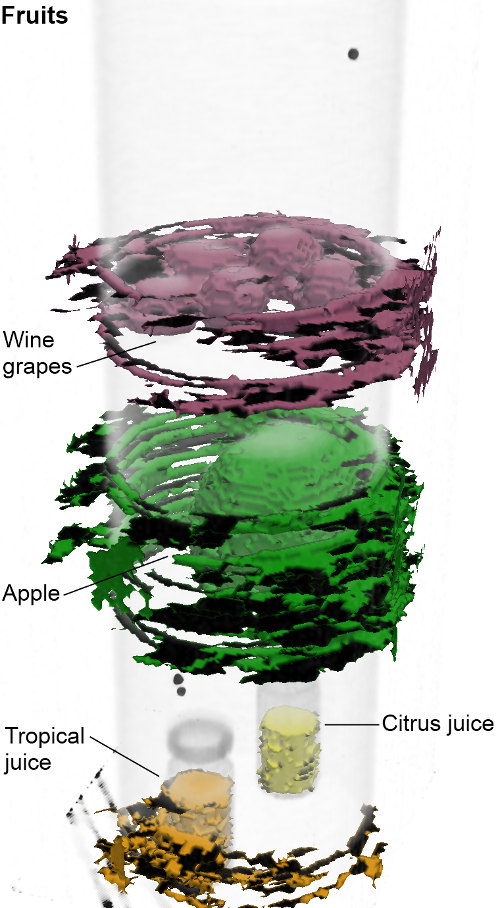}%
\label{fig:music3D:graphCut:Sample_06062018_Fruits}}
\subfloat{\includegraphics[width=0.245\linewidth,height=4.0cm,keepaspectratio]{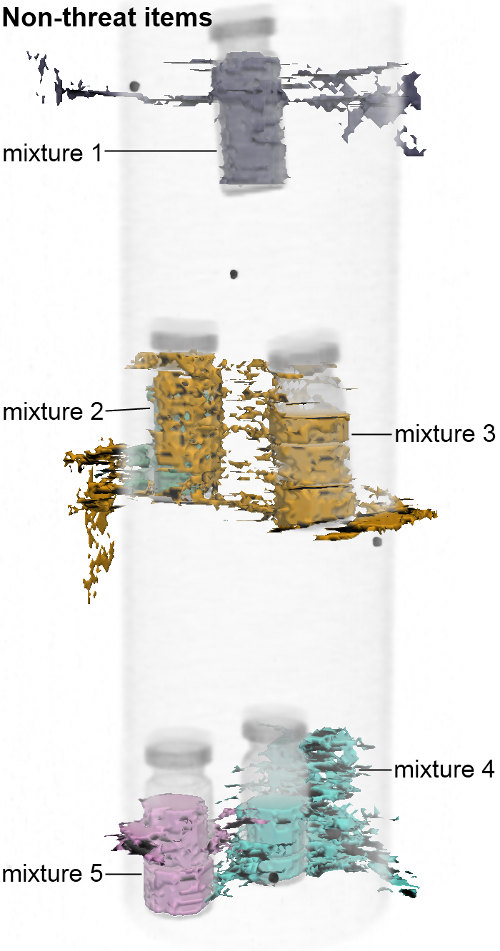}%
\label{fig:music3D:graphCut:Sample_06062018_NonThreat}}
\subfloat{\includegraphics[width=0.245\linewidth,height=4.0cm,keepaspectratio]{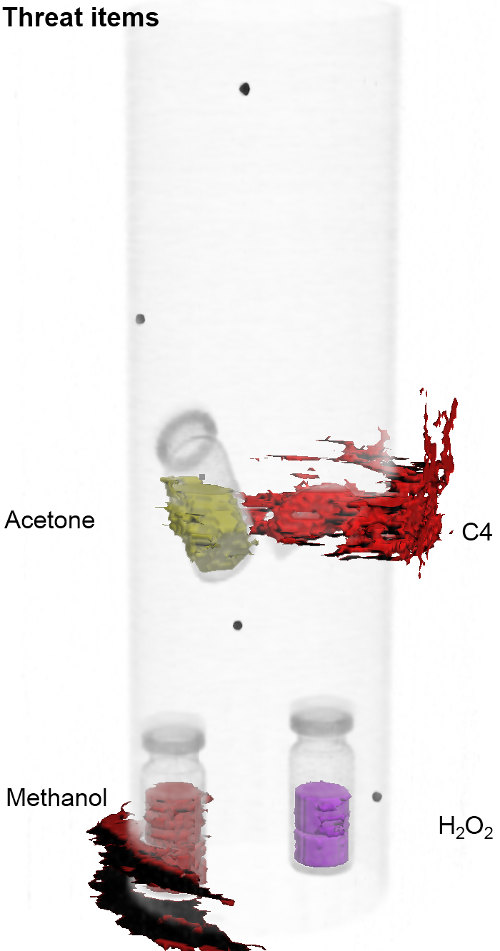}%
\label{fig:music3D:graphCut:Sample_06062018_Threat}}
\caption{Graph cut segmentation results achieved on the MUSIC3D dataset. Some segments are composed of multiple indicators, as in fig. \ref{fig:music2D:collageAttenuation:4samples}. Note the high degree of noise that leads to insufficient material distinction. The illustration embeds the segments (colours) in a DVR of energy channel 39 (61.35 keV).}
\label{fig:music3D:graphCut}
\end{figure}

The unconditioned graph cut also provides an improved segmentation quality for MUSIC3D, which is illustrated in fig. \ref{fig:music3D:graphCut}. As with \gls{FAMS}, due to the high noise level, the observation size parameter $k$ (see Felzenszwalb and Huttenlocher \cite{Felzenszwalb2004}) often needs to be adapted for each scan individually.

The overall segmentation quality of the graph cut is shown in table \ref{tab:music3D:dice} using the dice coefficient.

\begin{table}[h]
\caption{Dice coefficient overview for MUSIC3D dataset as quantitative segmentation quality evaluation}
\label{tab:music3D:dice}
\centering
\begin{tabular}{|l||r|}
\hline
\textbf{Sub-dataset} & \textbf{dice coeff.}\\
\hline
Sample 31102016 & 0.7646\\
\hline
Sample 23012018 & 0.3391\\
\hline
Sample 24012018 & 0.4424\\
\hline
Fluids & 0.3428\\
\hline
Fruits & 0.5578\\
\hline
Non-threat items & 0.6145\\
\hline
Threat items & 0.7010\\
\hline
\hline
Overall & 0.5374\\
\hline
\end{tabular}
\end{table}

\begin{figure}[htbp]
\centering
\subfloat{\includegraphics[width=0.325\linewidth]{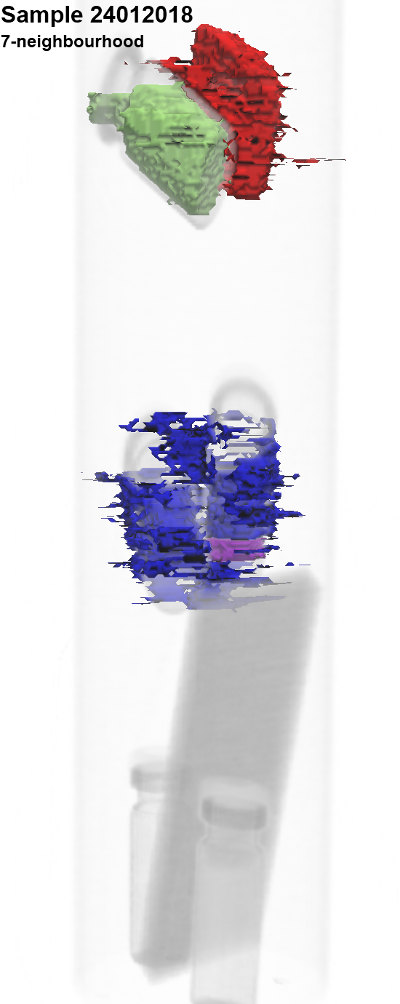}%
\label{fig:music3D:24012018:nbr:N7}}
\subfloat{\includegraphics[width=0.325\linewidth]{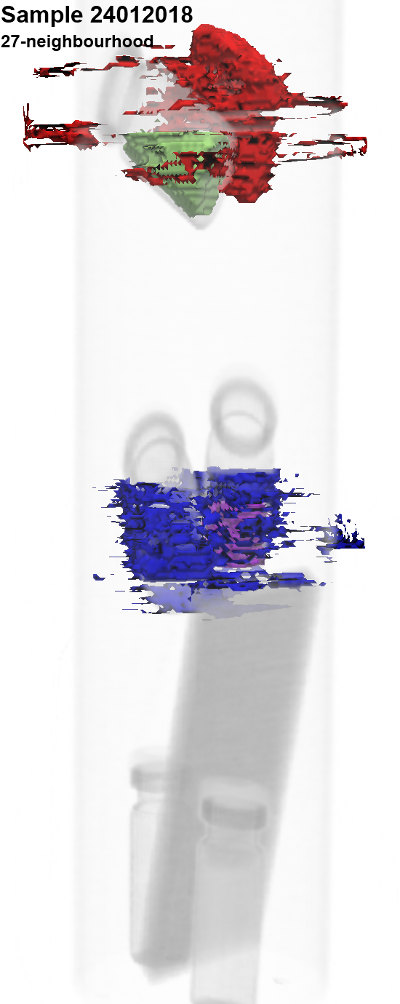}%
\label{fig:music3D:24012018:nbr:N27}}
\subfloat{\includegraphics[width=0.325\linewidth]{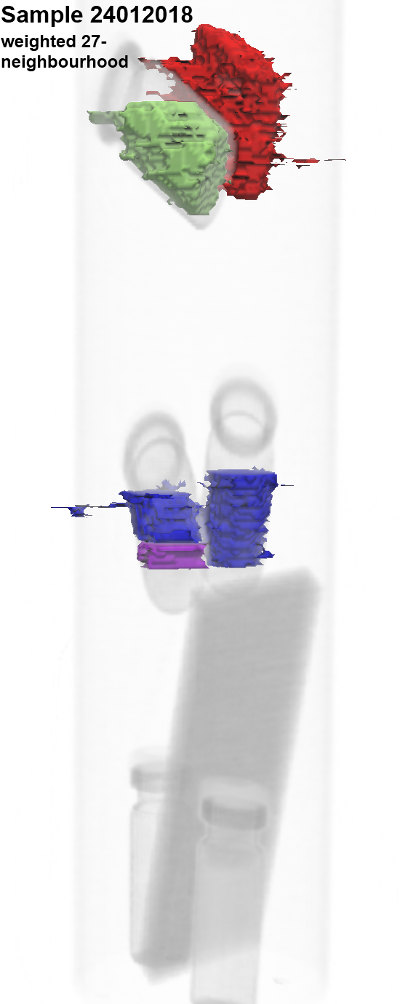}%
\label{fig:music3D:24012018:nbr:N27w}}
\caption{Comparison of the graph cut segmentation with full-spectrum data depending on different neighbourhood definitions. The illustration embeds the segments (colours) in a DVR of energy channel 39 (61.35 keV).}
\label{fig:music3D:24012018:nbr}
\end{figure}

The neighbourhood definition applied for the edge connectivity in graph $G(E,V)$, as described in sec. \ref{sec:algorithms:graphCut}, has a significant impact on the segmentation results. Fig. \ref{fig:music3D:24012018:nbr} shows the effect of varying neighbourhood definitions on the graph cut segmentation result: the common 27-neighbourhood definition performs relatively poorly as it is visually prone to noise. A 7-neighbourhood definition results in cleaner separations between segments, though at the cost of drastic oversegmentation (two to three times the overall extracted segments compared to other neighbourhood definitions). The weighted 27-neighbourhood definition gives the most visually-acceptable results for all full-spectrum 3D samples.

%

\subsection{Adaptive spectral binning}\label{sec:results:specBinning}

\begin{figure}[htb]
\begin{center}
  \centering
  \includegraphics[width=0.98\linewidth]{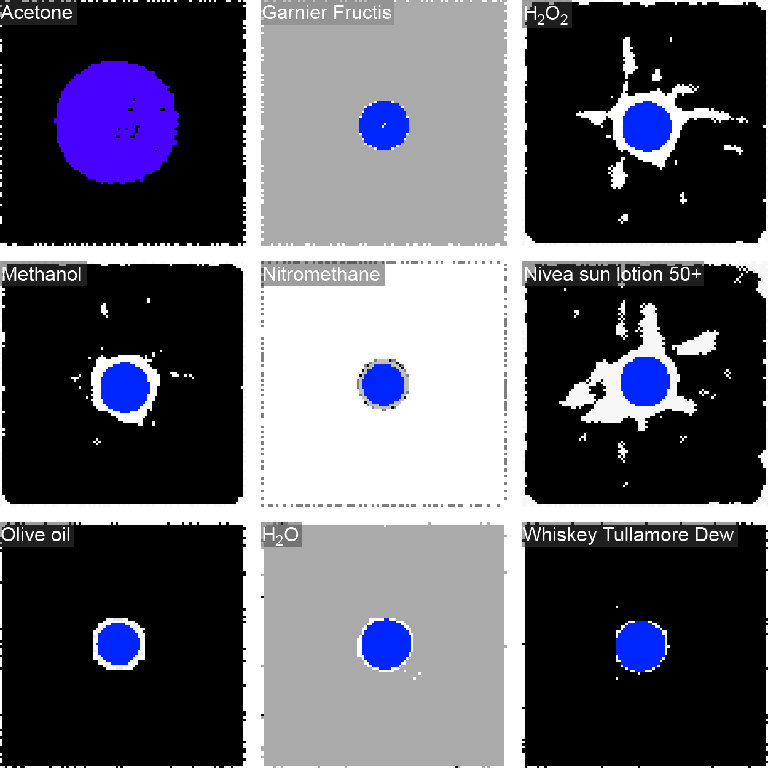}
  \caption{\label{fig:adaptiveBin:music2D:FAMS:refMats}Improvements of FAMS segmentation maps after adaptive binning.}
\end{center}
\end{figure}

\begin{figure}[htb]
\begin{center}
  \centering
  \includegraphics[width=0.98\linewidth]{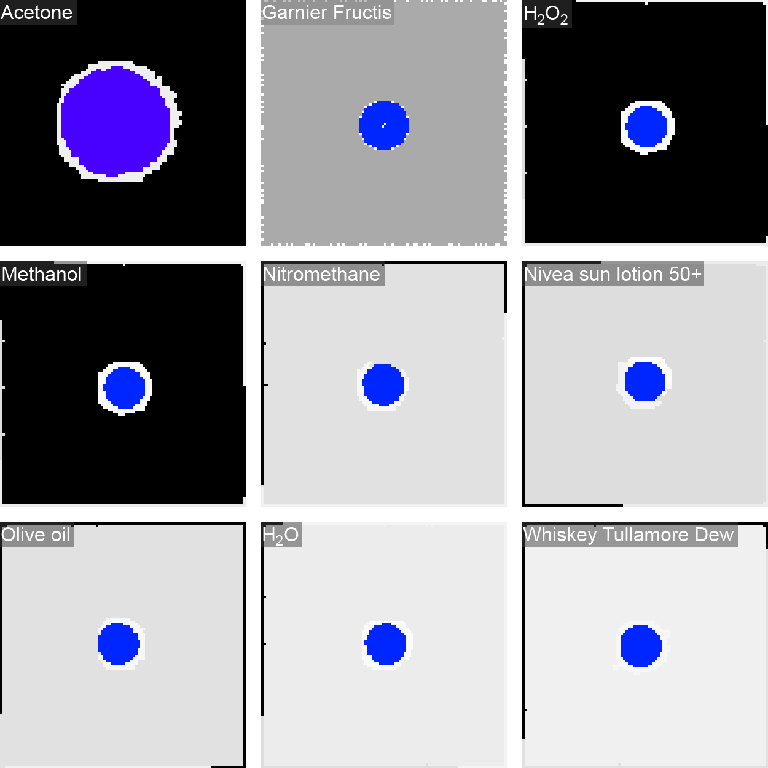}
  \caption{\label{fig:adaptiveBin:music2D:graphcut:refMats}Improvements of graph cut segmentation maps after adaptive binning.}
\end{center}
\end{figure}

\begin{figure}[htb]
\centering
\subfloat{\includegraphics[width=0.245\linewidth,height=4.1cm,keepaspectratio]{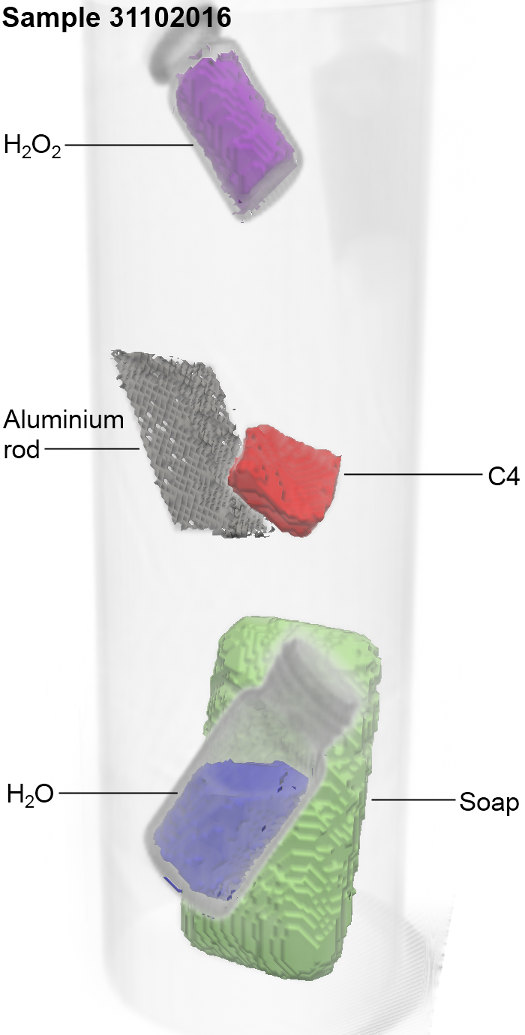}%
\label{fig:adaptiveBin:music3D:graphCut:Sample_31102016}}
\subfloat{\includegraphics[width=0.245\linewidth,height=4.1cm,keepaspectratio]{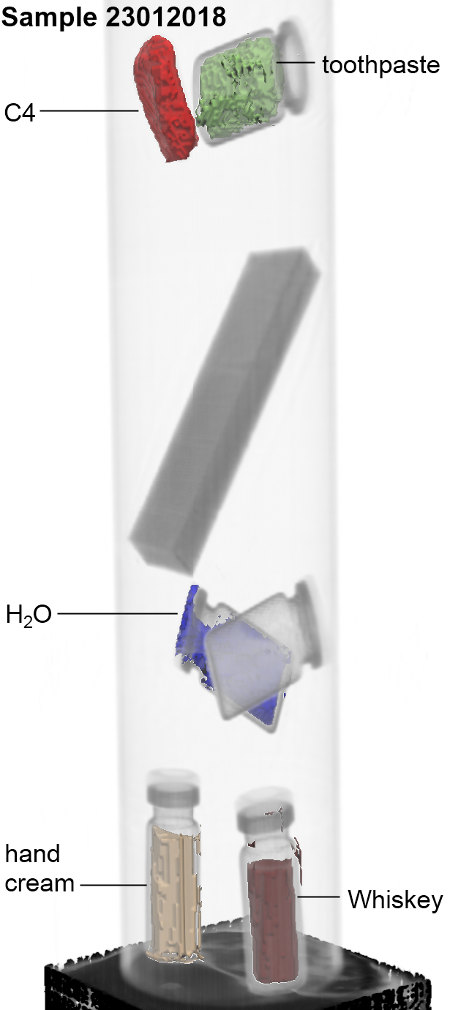}%
\label{fig:adaptiveBin:music3D:graphCut:Sample_23012018}}
\subfloat{\includegraphics[width=0.245\linewidth,height=4.1cm,keepaspectratio]{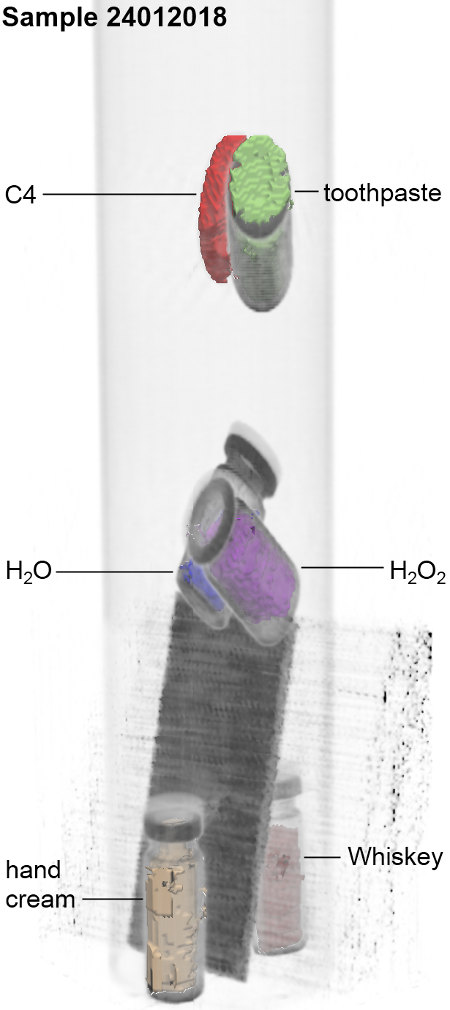}%
\label{fig:adaptiveBin:music3D:graphCut:Sample_24012018}}
\subfloat{\includegraphics[width=0.245\linewidth,height=4.1cm,keepaspectratio]{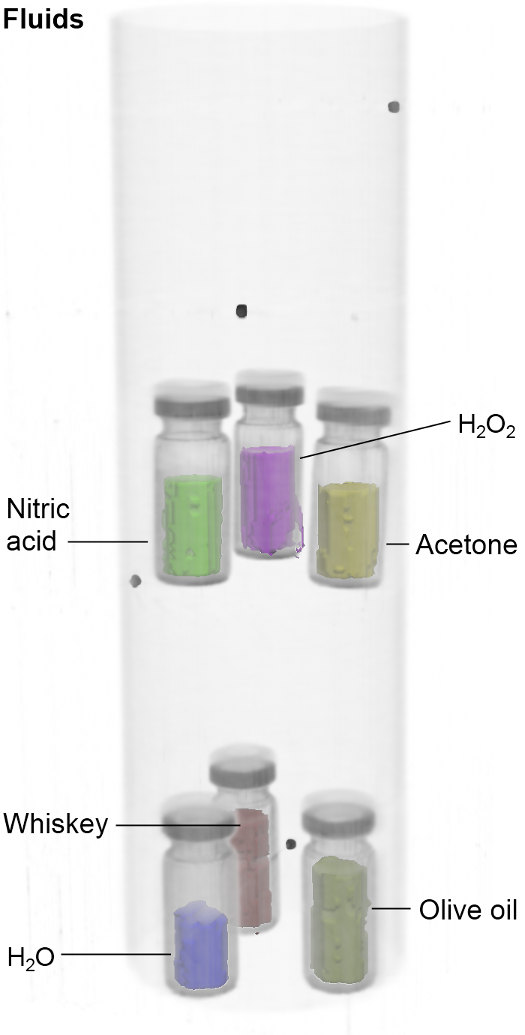}%
\label{fig:adaptiveBin:music3D:graphCut:Sample_06062018_Fluids}}
\hfil
\subfloat{\includegraphics[width=0.245\linewidth,height=4.1cm,keepaspectratio]{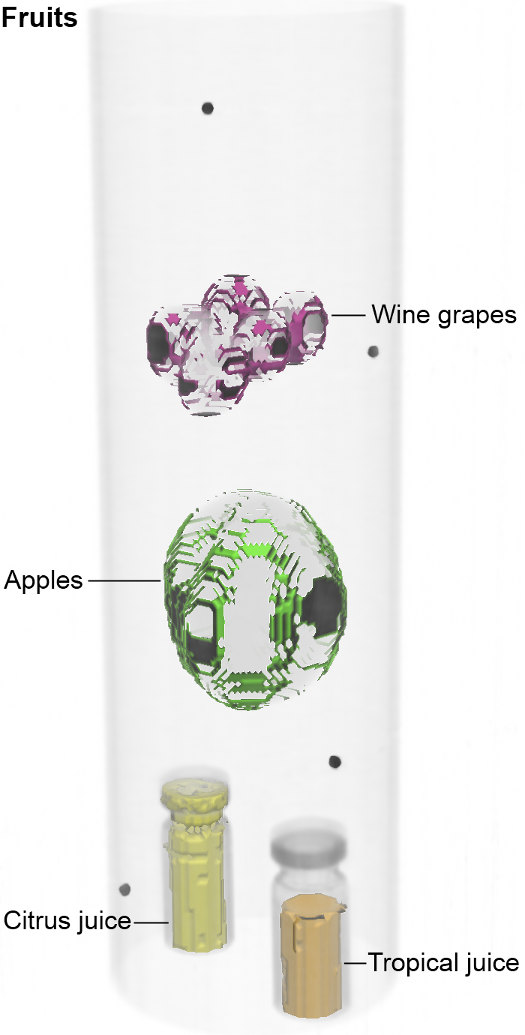}%
\label{fig:adaptiveBin:music3D:graphCut:Sample_06062018_Fruits}}
\subfloat{\includegraphics[width=0.245\linewidth,height=4.1cm,keepaspectratio]{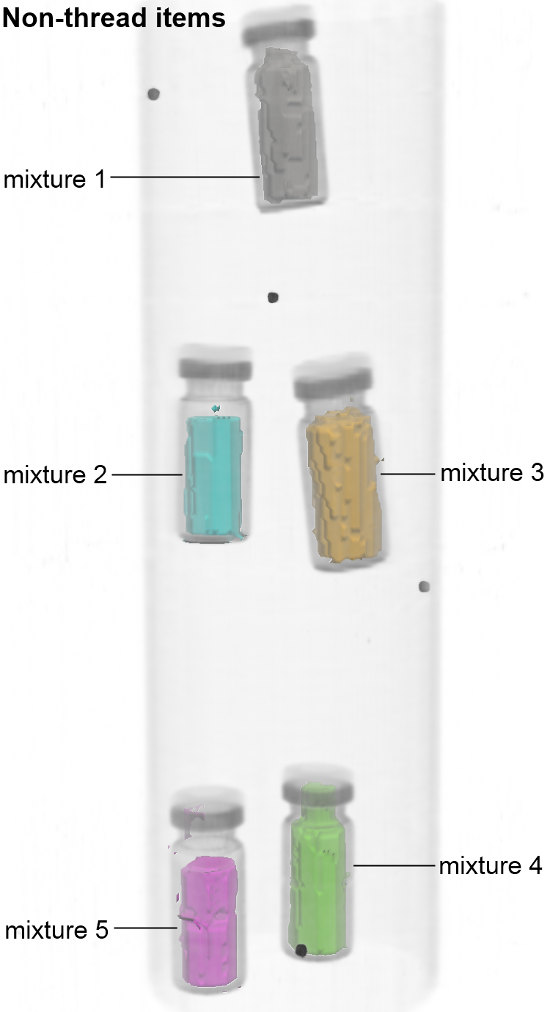}%
\label{fig:adaptiveBin:music3D:graphCut:Sample_06062018_NonThreat}}
\subfloat{\includegraphics[width=0.245\linewidth,height=4.1cm,keepaspectratio]{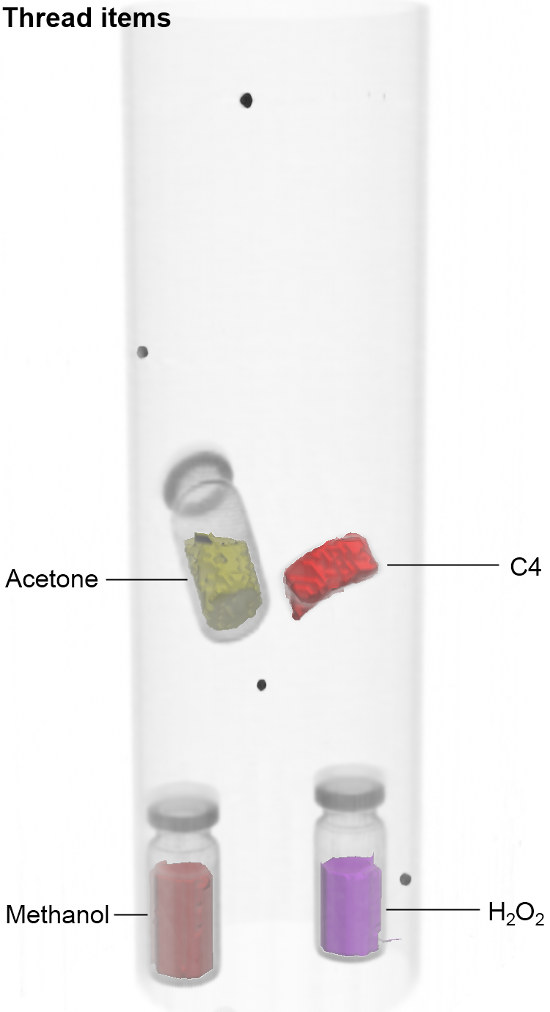}%
\label{fig:adaptiveBin:music3D:graphCut:Sample_06062018_Threat}}
\caption{MUSIC3D graph cut segmentation results achieved after adaptive binning. The illustration embeds the segments (colours) in a DVR of energy channel 39 (61.35 keV).}
\label{fig:adaptiveBin:music3D:graphCut}
\end{figure}

The application of the adaptive, anisotropic spectral binning scheme generally offers large improvements upon the poor segmentations presented until now. The improved segmentation quality can be observed quantitatively in the dice coefficient (see table \ref{tab:adaptiveBin:music2D:dice} and \ref{tab:adaptiveBin:music3D:dice}) as well as qualitatively by visual inspection (see fig. \ref{fig:adaptiveBin:music2D:FAMS:refMats}, \ref{fig:adaptiveBin:music2D:graphcut:refMats} and \ref{fig:adaptiveBin:music3D:graphCut}).


\begin{table}[h]
\caption{Dice coefficient overview of the graph cut for MUSIC2D dataset after adaptive binning}
\label{tab:adaptiveBin:music2D:dice}
\centering
\begin{tabular}{|l||r|}
\hline
\textbf{Sub-dataset} & \textbf{dice coeff.}\\
\hline
Acetone            & 0.4340\\
\hline
Brandy Chantr\'{e} & 0.6613\\
\hline
Cien hand cream    & 0.8676\\
\hline
Garnier Fructis    & 0.8961\\
\hline
H$_2$O$_2$         & 0.6749\\
\hline
Methanol           & 0.9155\\
\hline
Nitromethane       & 0.4545\\
\hline
Nivea sun lotion 50+ & 0.8301\\
\hline
Olive oil          & 0.4778\\
\hline
H$_2$O             & 0.4992\\
\hline
Whiskey Tullamore Dew & 0.5011\\
\hline
\hline
Overall            & 0.6557\\
\hline
\end{tabular}
\end{table}

\begin{table}[h]
\caption{Dice coefficient overview for MUSIC3D dataset after adaptive binning}
\label{tab:adaptiveBin:music3D:dice}
\centering
\begin{tabular}{|l||r|}
\hline
\textbf{Sub-dataset} & \textbf{dice coeff.}\\
\hline
Sample 31102016 & 0.7630\\
\hline
Sample 23012018 & 0.5155\\
\hline
Sample 24012018 & 0.8758\\
\hline
Fluids & 0.9095\\
\hline
Fruits & 0.4195\\
\hline
Non-threat items & 0.8995\\
\hline
Threat items & 0.7569\\
\hline
\hline
Overall & 0.7342\\
\hline
\end{tabular}
\end{table}

Adaptive binning additionally allows for determining a fixed observation size parameterisation for \gls{FAMS} and graph cuts that is valid across all scans, discarding the need for manual parameter optimisation. This is due to the reduced noise in each target energy bin, the increase in self-information carried by each energy channel, and the data-adaptive nature of the binning. Thus, the obtained results for adaptive binning use the following parameterisation:

\textbf{FAMS}:
\begin{itemize}
\item k: 220
\item K: 24
\item L: 35
\end{itemize}

\textbf{unconditioned graph cuts}:
\begin{itemize}
\item k: 3.0
\item minSize: 625
\item edge connectivity: 27-neighbourhood
\end{itemize}

\begin{figure}[htbp]
\centering
\subfloat{\includegraphics[width=0.24\linewidth]{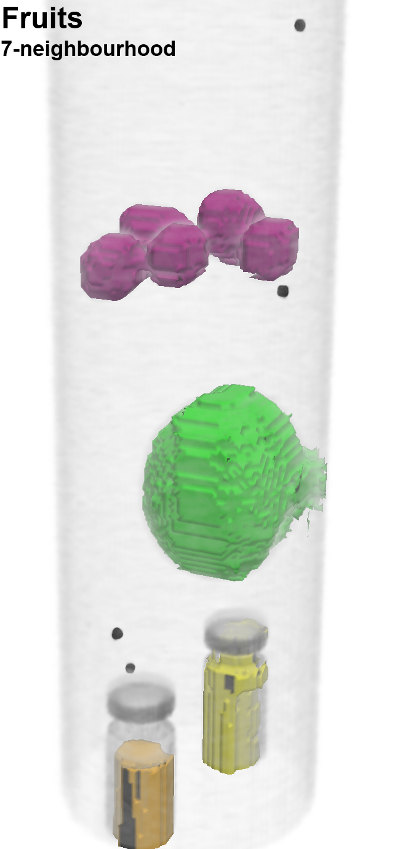}%
\label{fig:adaptiveBin:music3D:nbr:N7}}
\subfloat{\includegraphics[width=0.24\linewidth]{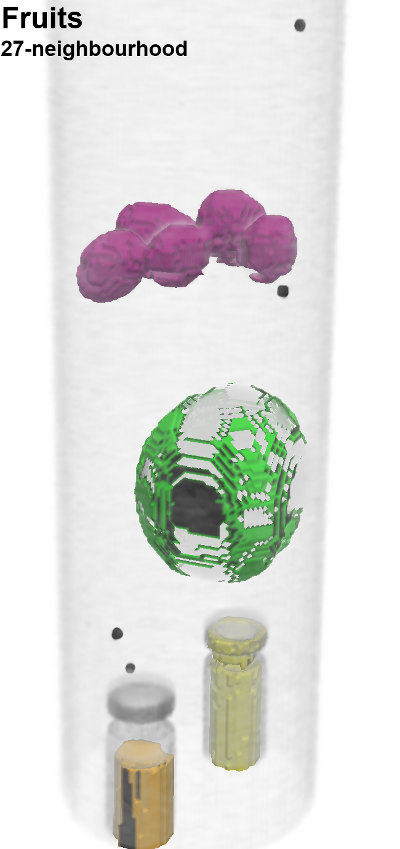}%
\label{fig:adaptiveBin:music3D:nbr:N27}}
\subfloat{\includegraphics[width=0.24\linewidth]{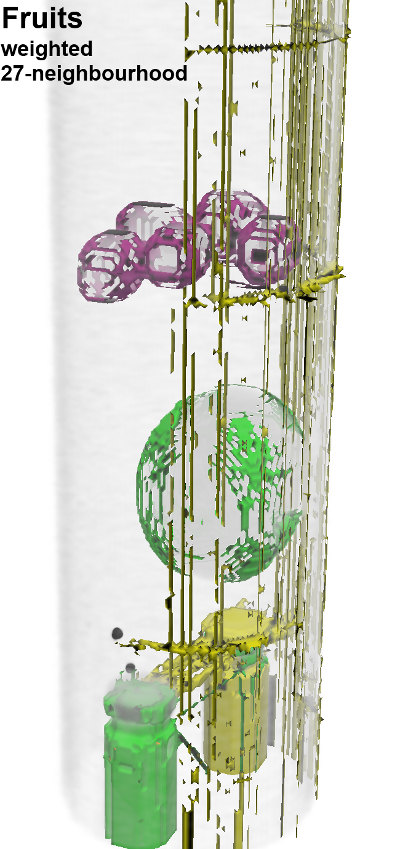}%
\label{fig:adaptiveBin:music3D:nbr:N27w}}
\caption{Comparison of the graph cut segmentation with adaptive binning depending on different neighbourhood definitions. The illustration embeds the segments (colours) in a DVR of energy channel 39 (61.35 keV).}
\label{fig:adaptiveBin:music3D:nbr}
\end{figure}

The type of neighbourhood definition has a distinct impact on segmentation results by the graph cut method, thus fig. \ref{fig:adaptiveBin:music3D:nbr} shows the achievable quality of different neighbourhood definitions on adaptively binned data: within the 'Fruits' sample, the objects are located closely to the bounding plexi-glass cylinder. Due to the lack of a definite hull separating the plexi-glass and the objects, leakage of the segments occurs so that fruits and plexi-glass are incorrectly labelled with an equal indicator. This leakage can be prevented when applying a 7-neighbourhood definition. This is, in cases of cluttered object arrangements such as check-in luggage, a good control to steer the target level of detail. Note that this result on adaptively-binned data is in contrast to full-spectrum graph cut segmentation and that the weighted 27-neighbourhood definition performs poorly with adaptive binning. The reason for the contrasting results is in the improved \gls{SNR} for adaptively-binned data: a 7-neighbourhood kernel on full-spectrum data extracts an excessive amount of segments that are poorly connected between slices in the volume stack due to high noise, which is filtered with the weighted 27-neighbourhood kernel. For the case of adaptive binning, the improved \gls{SNR} allows for better-connected segments in general, which explains the superior 7-neighbourhood kernel performance.

\subsection{CT reconstruction influence}\label{sec:results:ctReconstruction}

When comparing scan 'Sample 31102016' and 'Sample 24102016' in MUSIC3D (see fig. \ref{fig:music3D:collageAttenuation}), we observed a difference in segmentation quality due to metal artefact influence. The metal artefacts are partially so dominant that material identification is impossible. \gls{MAR} is needed and, in common cases with metal artefacts only affecting minor portions of the whole scan, it can possibly eliminate the metalicity issue within scans for material identification \cite{Barrett2004}. In other application areas, \gls{MAR} is still a problem with with recent advances \cite{Karimi2015} that requires further treatment in the literature\cite{Bongers2015}.

\begin{figure}[htbp]
\centering
\subfloat{\includegraphics[width=0.24\linewidth]{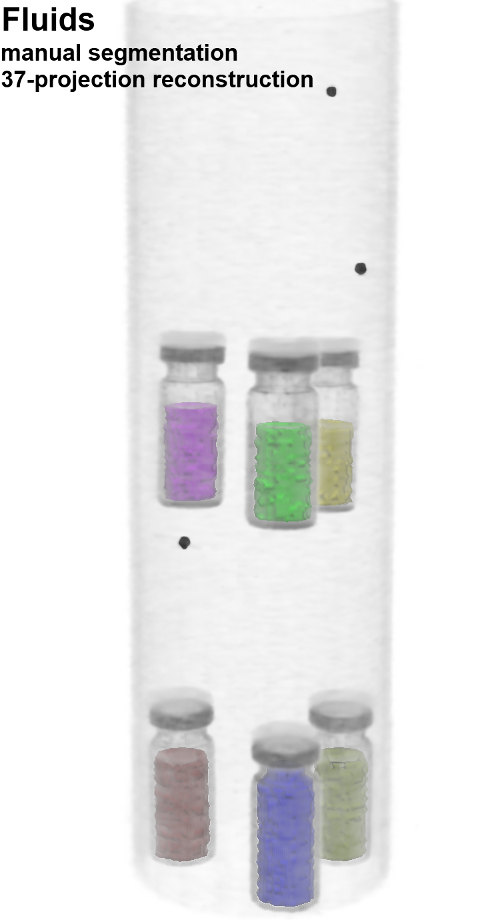}%
\label{fig:adaptiveBin:music3D:projections:manual}}
\subfloat{\includegraphics[width=0.24\linewidth]{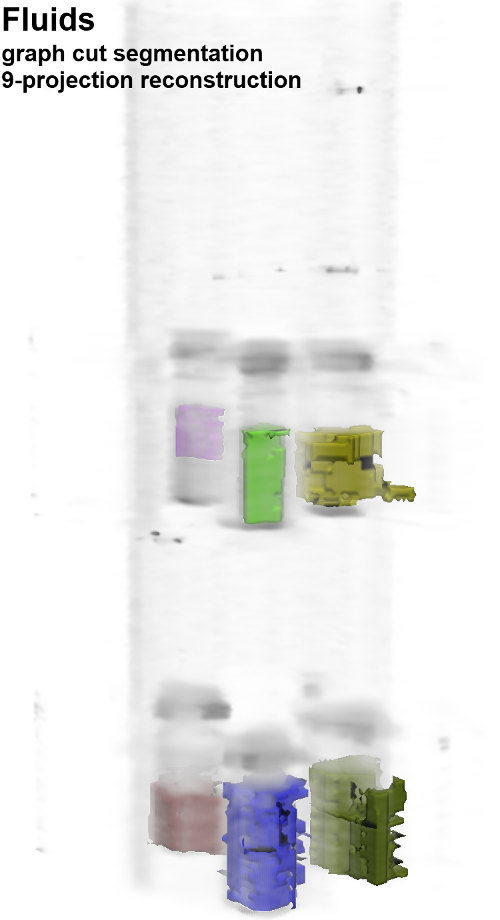}%
\label{fig:adaptiveBin:music3D:projections:proj9}}
\subfloat{\includegraphics[width=0.24\linewidth]{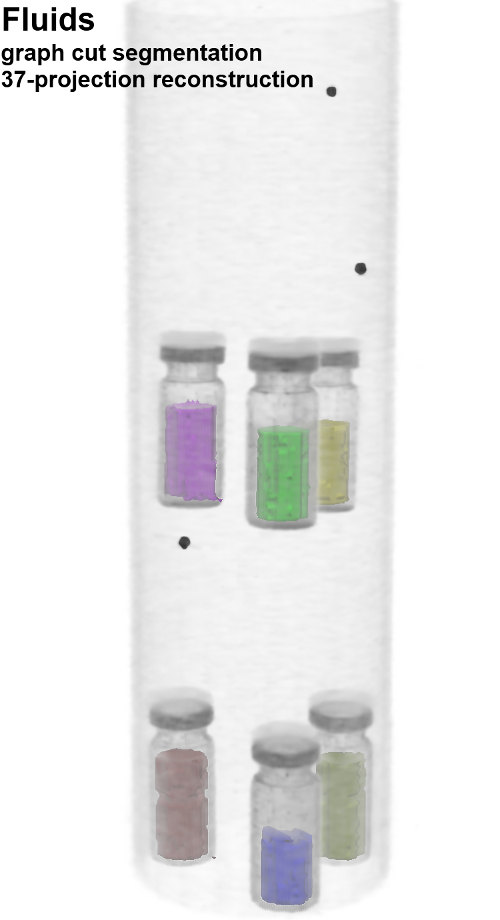}%
\label{fig:adaptiveBin:music3D:projections:proj37}}
\subfloat{\includegraphics[width=0.24\linewidth]{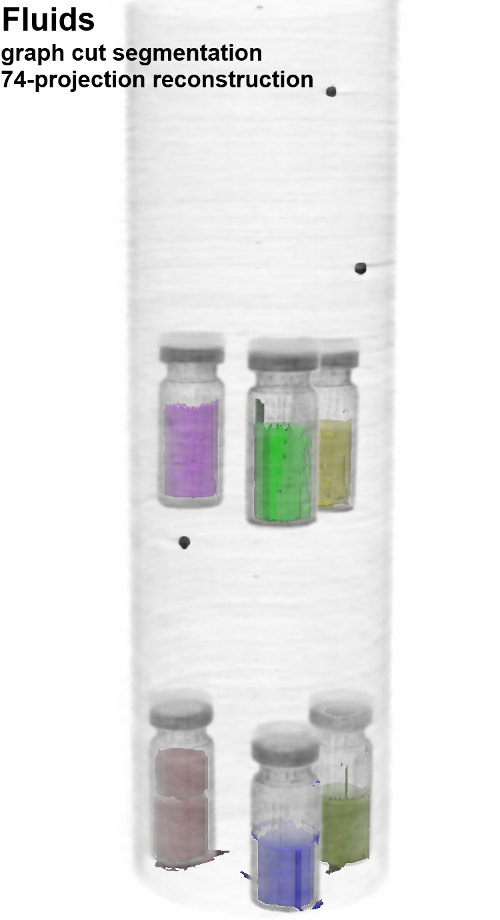}%
\label{fig:adaptiveBin:music3D:projections:proj74}}
\caption{Comparison of the graph cut segmentation with adaptive binning depending on different number of projections used for the reconstruction with the reference manual segmentation. The illustration embeds the segments (colours) in a DVR of energy channel 39 (61.35 keV).}
\label{fig:adaptiveBin:music3D:projections}
\end{figure}

Another significant influence on \gls{SNR} and \gls{CNR} is rooted in the inverse reconstruction from x-ray projections to the computed tomography, as discussed in section \ref{sec:literature:reconstruction}. With the projections available from compressed sensing, an analytical reconstruction for the \gls{MECT} data is not possible. In an undersampled tomography domain, image quality (with respect to \gls{SNR} and \gls{CNR}) degrades rapidly with a decrease in available projection data. Fig. \ref{fig:adaptiveBin:music3D:projections} shows the effects of this decreasing image quality with the availability of 74, 37 and 9 projections on the segmentation using the above-outlined adaptive binning and unconstrained graph cut. As can be seen in the images, even using a considerably undersampled tomographic scan (9 projections), which leads to severe reconstruction artefacts, the presented segmentation procedure allows to extract a reasonable segmentation.

%


\section{Conclusion}

In conclusion, we presented an openly available spectral \gls{CT} dataset, acquired in the domain of baggage scanning, which is aimed at improving automatic image analysis by method benchmarking. The dataset itself, called \textit{MUSIC}, consists of two separate parts for 2D (32 sets)- and 3D (7 sets) x-ray projections and tomographic reconstructions. The dataset, including further information on how to use it, is available at \blue{\url{http://easi-cil.compute.dtu.dk/index.php/datasets/music/}}. The set includes the corrected x-ray projections, their \gls{ART-TV} reconstructions, as well as the presented segmentations (for FAMS and graph cuts). The data are provided in the following formats:

\begin{itemize}
\item MatLab/Octave/Python for 3D data: HDF5 (.h5)
\item C++/Python/available CT software: MetaIO format (.mhd)
\end{itemize}

Furthermore, the C++ implementation of the unconditioned graph cuts as well as the utilized Python scripts in the data processing can be obtained at \blue{\url{https://www.github.com/CKehl/MECT.git}}.

As for the methodology analysis, we presented and compared two techniques for fully unsupervised spectral image- and volume segmentation, namely 3D spectral \gls{FAMS} and unconditioned 3D spectral graph cuts, that are based on existing literature and that were adapted to process \gls{MECT} data. The article compared the results of both segmentation methods using an isotropic, uniform spectral binning from 128 down to 20 energy channels of the \textit{MUSIC} dataset, which shows significant drawbacks of both methods in the presence of high noise and low \glspl{SNR}. Based on the results for isotropic binning, we presented and adaptive, anisotropic spectral binning that follows a variance budget allocation scheme. The adaptive binning scheme, using 10 compressed energy bins, improves the \gls{SNR} per energy bin while maintaining the information diversity in low-energy channels. The improvements are observable visually in the segmentation volume maps as well as quantitatively in the dice coefficient measurements ($\text{dice}_{coeff}=0.7234$ for 2D spectral-, $\text{dice}_{coeff}=0.3622$ for 3D spectral data). Besides the improves \gls{SNR}, the adaptive binning also eliminates the need for meticulous manually-tuned parameterisation of each segmentation method. 

Evaluating the segmentation methods in more detail, we generally observe better segmentation maps given by the unconditioned graph cut while the \gls{MS}-based algorithm tends to supply object boundary maps (similar to edge filtering). The result of the \gls{FAMS} method on \gls{MECT} data shows significantly different behaviour than previously-published results on hyperspectral imaging in the visible- and infrared part of the spectrum \cite{Jordan2013}. A detailed analysis has revealed that the different results may be rooted in the problem of signal quantization: the \gls{FAMS} method was designed for images obtained by digital cameras with common \gls{CCD} sensors and optical lens filters to separate the various spectra, quantifying the recorded light intensities in 8- or 16-bit values (irrespective of integer- or floating-point value representation). For tomographic reconstructions and with the goal to separate marginal material differences, a 16-bit quantization range is insufficient (see fig. \ref{fig:music2D:LAC} to get an impression of the recorded attenuation range and the quantization resolution required to differentiate various fluids). In our experiments, the spectral gradient itself is significant enough to detect material boundaries, but the underlying signal response (i.e. the \gls{LAC}) does not facilitate multi-material differentiation. As discussed in section \ref{sec:results:fams}, a connected-component analysis based on clean object separation via \gls{FAMS} may yield appropriate segmentation maps for later processing.

Lastly, this article illustrated and discussed deteriorating affects, such as metal artefacts and the considerable undersampling of x-ray projections in \gls{ART-TV} reconstruction, on the segmentation quality. Both of these issues are still remaining obstacles that need to be addressed in future work on the subject. It is particularly important for the application of automatic check-in luggage scanning and material identification because (i) metal objects cannot be excluded from the luggage and (ii) the physical constraints of airport \gls{CT} scanners in some cases do not allow for more than 9 x-ray detectors (i.e. 9 x-ray projections) being acquired simultaneously.

\section{Discussion}

More high-quality segmentations may be achievable with other existing methods. The strongest postulate of the presented research is the unknown number of objects and materials within a specific dataset. We apply this postulate to automatically-acquired scans because, from a strict statistical perspective, predefining the number of expected segments introduces a \textit{bias} in the segmentation. This bias is variable and depends on the observation scale, the segment definition (or definition policy) or the personal judgement of the domain expert with respect to the distinction detail. As such, the bias allows for flexibility and uncertainty in the segmentation itself and provides an error margin for initial estimations. \textit{A priori} knowledge about the number of distinct materials or even their approximate position inside the scan (i.e. material seeds) facilitates for more robust semi-supervised segmentations (e.g. minCut-maxFlow graph cuts, gaussian mixture models). Hence, our future research involves the robust, probabilistic estimation of material seed points inside \gls{MECT} scans.

The unsupervised methods presented here deliver numerous equiprobable segmentation for each scan, which provide large reference datasets for subsequent machine learning approaches (e.g. neural networks). The increasing amount of segmentation volumes enables the training of neural networks from a limited number of reference scans. One of the potentially major uses of \textit{MUSIC} is thus the provision of a spectral volume image database for also other applications that require material classification without shape priors.

\ifCLASSOPTIONcompsoc
  \section*{Acknowledgments}
\else
  \section*{Acknowledgment}
\fi

The authors would like to thank Mark Lyksborg and Mina Kheirabadi for their former MECT research, on which some concepts in this paper extent and build upon. We further acknowledge the Innovation Fund Denmark, funding the presented research in the CIL2018 project (code: 10437). We further thank and acknowledge NVIDIA Corp. for the donation of one NVIDIA Titan Xp in support of our research efforts.

\ifCLASSOPTIONcaptionsoff
  \newpage
\fi



\bibliographystyle{IEEEtran}
\bibliography{literature_TransImgProc}

%

\newpage
\begin{IEEEbiography}[{\includegraphics[width=1in,height=1.25in,clip,keepaspectratio]{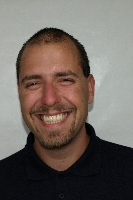}}]{Christian Kehl}
is a postdoctoral fellow at DTU Compute (Image Analysis and Computer Graphics section) since 2017. Christian received his Ph.D. in Mathematics and Natural Sciences in 2017 from the University of Bergen (Norway). He had former research affiliations with TU Delft (Computer Graphics and Visualization group, NL), University of Amsterdam (Institute of Informatics, NL), Uni Research AS Bergen (CIPR, NO) and Aix-Marseille Universit\'{e} (FR). His research interests are in high-performance image analysis, scientific data visualisation and discrete geometry algorithms and data structures.
\end{IEEEbiography}

\begin{IEEEbiography}
[{\includegraphics[width=1in,height=1.25in,clip,keepaspectratio]{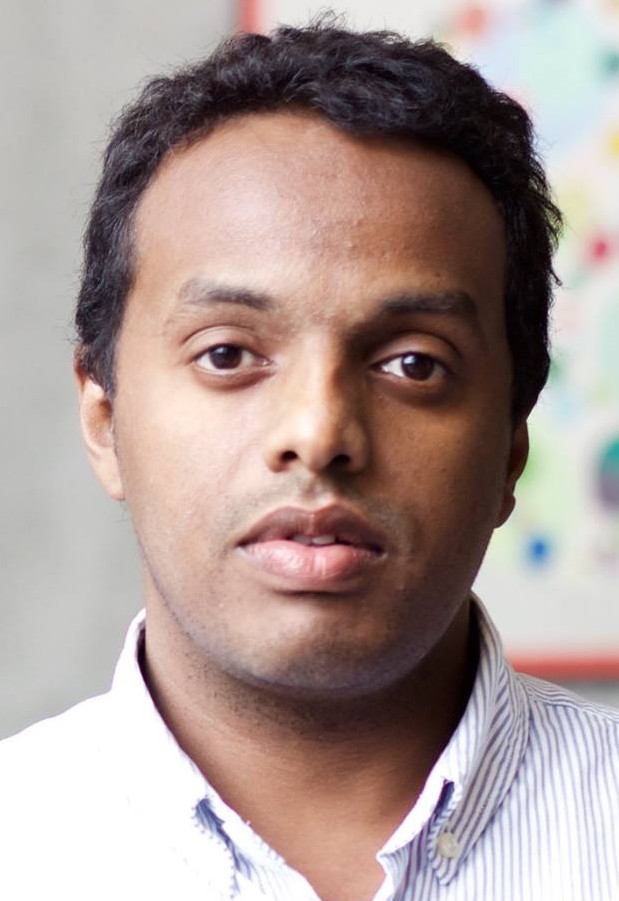}}]{Wail Mustafa}
received his B.Sc degree in electrical engineering from University of Khartoum, Sudan, in 2006, his M.Sc. degree in electrical engineering (signal processing) from Blekinge Institute of Technology, Sweden, in 2010, and his PhD degree in robotics (computer vision) from University of Southern Denmark in 2015. Since 2017, he has been a postdoctoral researcher at The Technical University of Denmark (DTU). His main research interests are computer vision and machine learning for cognitive and autonomous systems.
\end{IEEEbiography}

\begin{IEEEbiography}[{\includegraphics[width=1in,height=1.25in,clip,keepaspectratio]{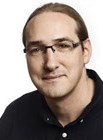}}]{Jan Kehres}
is a research engineer at Technical University of Denmark (DTU) Physics, Kongens Lyngby, Denmark. He received his PhD in operando investigation of catalyst nanoparticles using x-ray scattering and continued in this field of research as a postdoctoral researcher. His current field of research is the application of energy-dispersive detectors for x-ray scattering and advanced imaging modalities with a focus on material identification of illicit materials for security applications.
\end{IEEEbiography}

\begin{IEEEbiography}[{\includegraphics[width=1in,height=1.25in,clip,keepaspectratio]{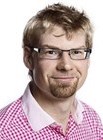}}]{Ulrik Lund Olsen}
is a Sr. research engineer at DTU physics since 2013. Funded by Innovation Fund Denmark to develop applications using high flux multispectral x-ray detection technology and currently project leader on CIL2018, a multidisciplinary effort made to reduce the human operator involvement by 50\% for checked-in luggage. He was previously employed at Risø National Laboratory, as PhD (2009) and later postdoc working with developments of x-ray sensors.
\end{IEEEbiography}

\begin{IEEEbiography}[{\includegraphics[width=1in,height=1.25in,clip,keepaspectratio]{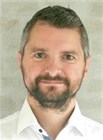}}]{Anders Bjorholm Dahl}
Biography text here.
\end{IEEEbiography}







\newpage
\appendices

\section{2D multi-object composition}
\label{app:multi_obj_compose}

\begin{figure}[htbp]
\begin{center}
	\subfloat{\includegraphics[width=0.49\linewidth,height=4.2cm,keepaspectratio]{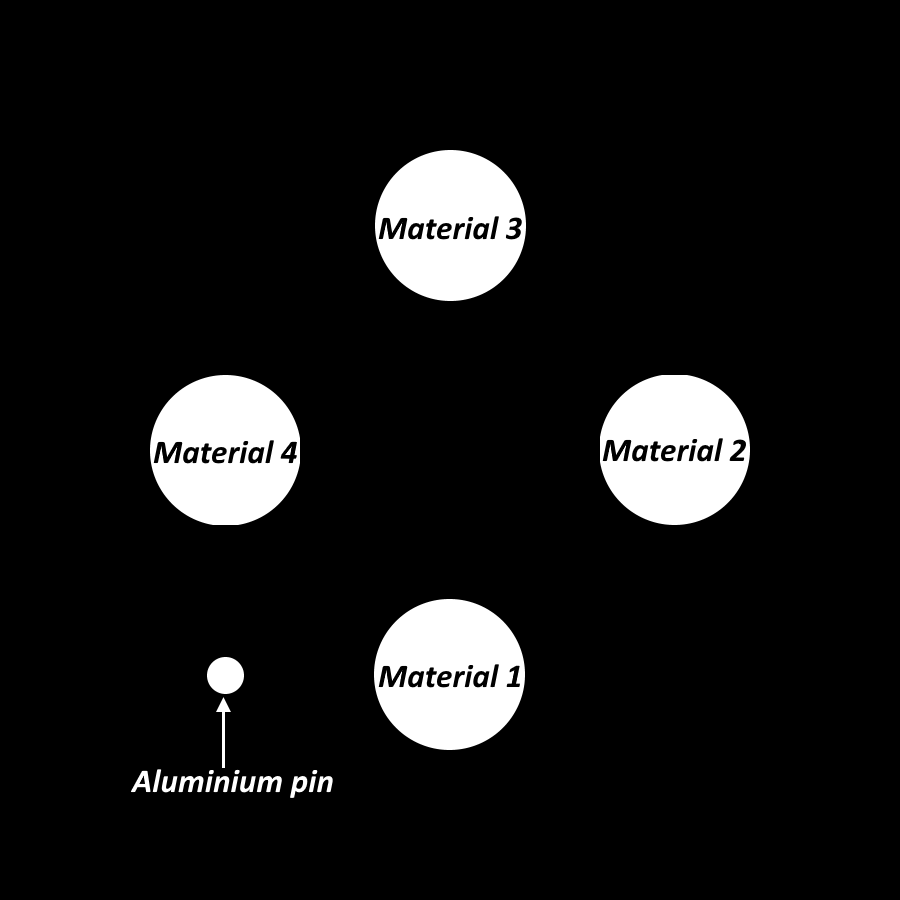} \label{fig:music2D:layout:centered}}
	\subfloat{\includegraphics[width=0.49\linewidth,height=4.2cm,keepaspectratio]{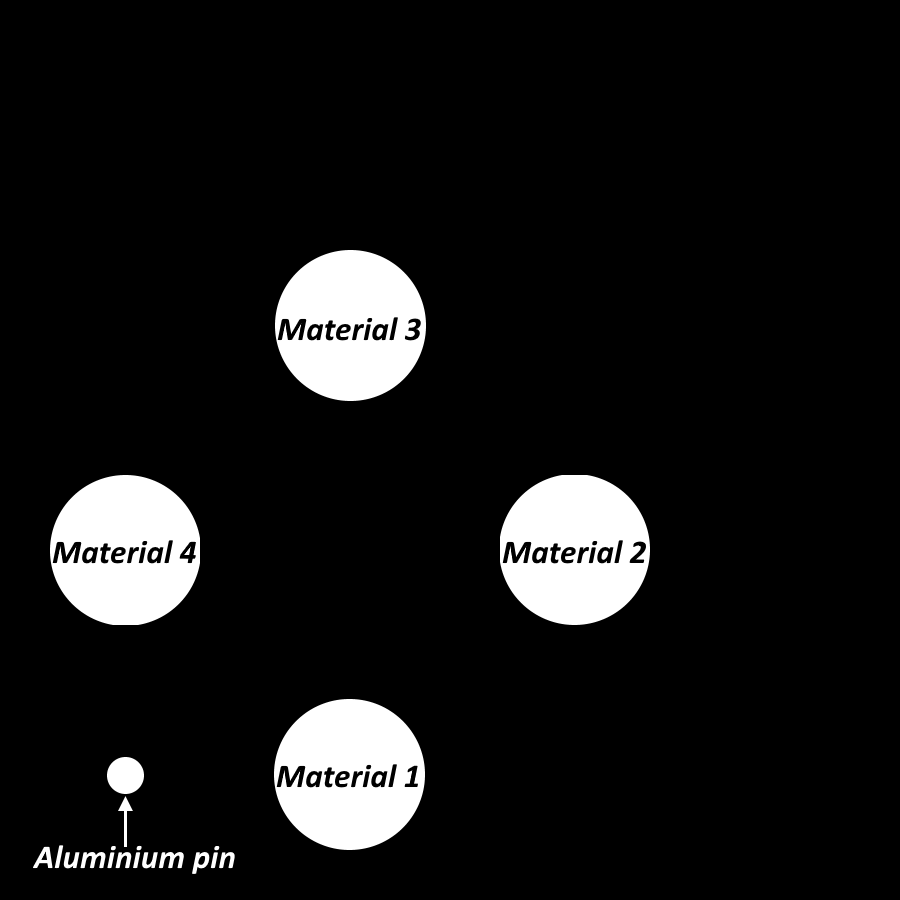} \label{fig:music2D:layout:noncentered}}
  \caption{\label{fig:music2D:layout} Sketch on the location of each sample referred to in tab. \ref{tab:music2D:composition_mats}.}
\end{center}
\end{figure}

\begin{table*}[h]
\caption{Complete overview of all materials inside the composed MUSIC2D subset. The naming is counter-clockwise, starting at the top. An aluminium pin at top-right hand corner and coordinate system per image originates top-left hand corner.}
\label{tab:music2D:composition_mats}
\centering
\begin{tabular}{|l||r|r|r|r|}
\hline
\textbf{Sub-dataset} & \textbf{Material 1} & \textbf{Material 2} & \textbf{Material 3} & \textbf{Material 4}\\
\hline
Sample 1 & Olive oil & Brandy Chantr\'{e} & Cien hand cream & Whiskey Tullamore Dew \\
\hline
Sample 2 & Nivea sun lotion 50+ & Brandy Chantr\'{e} & Methanol & H$_2$O$_2$ (50\%) \\
\hline
Sample 3 & Garnier Fructis & Whiskey Tullamore Dew & Nitromethane & Brandy Chantr\'{e} \\
\hline
Sample 4 & H$_2$O & H$_2$O$_2$ (50\%) & Garnier Fructis & Nivea sun lotion 50+ \\
\hline
Sample 5 & Cien hand cream & Nitromethane & Whiskey Tullamore Dew & H$_2$O \\
\hline
Sample 6 & H$_2$O & Brandy Chantr\'{e} & Olive oil & Whiskey Tullamore Dew \\
\hline
Sample 7 & Brandy Chantr\'{e} & Garnier Fructis & H$_2$O$_2$ (50\%) & Nivea sun lotion 50+ \\
\hline
Sample 8 & H$_2$O$_2$ (50\%) & Methanol & Garnier Fructis & Olive oil \\
\hline
Sample 9 & Methanol & Cien hand cream & Acetone & Nivea sun lotion 50+ \\
\hline
Sample 10 & Acetone & H$_2$O & Garnier Fructis & Whiskey Tullamore Dew \\
\hline
Sample 11 & Acetone & H$_2$O & Garnier Fructis & Whiskey Tullamore Dew \\
\hline
Sample 12 & Brandy Chantr\'{e} & Olive oil & Methanol & H$_2$O$_2$ (50\%) \\
\hline
Sample 13 & Garnier Fructis & Whiskey Tullamore Dew & H$_2$O & Acetone \\
\hline
Sample 14 & Nitromethane & Nivea sun lotion 50+ & Cien hand cream & Brandy Chantr\'{e} \\
\hline
Sample 15 & H$_2$O$_2$ (50\%) & Olive oil & Garnier Fructis & Nitromethane \\
\hline
Sample 16 & Nivea sun lotion 50+ & Whiskey Tullamore Dew & H$_2$O & Olive oil \\
\hline
Sample 17 & Whiskey Tullamore Dew & Brandy Chantr\'{e} & Garnier Fructis & Nitromethane \\
\hline
Sample 18 & Cien hand cream & Nivea sun lotion 50+ & Acetone & H$_2$O$_2$ (50\%) \\
\hline
Sample 19 & Olive oil & Garnier Fructis & Cien hand cream & Brandy Chantr\'{e} \\
\hline
Sample 20 & H$_2$O & Acetone & Olive oil & Whiskey Tullamore Dew \\
\hline
Sample Test & Olive oil & Brandy Chantr\'{e} & Whiskey Tullamore Dew & Cien hand cream \\
\hline
\end{tabular}
\end{table*}

\section{3D manual segmentations}
\label{app:manualSegmentations3D}

\begin{figure}[htbp]
\begin{center}
  \centering
  \includegraphics[width=0.75\linewidth,height=12.1cm,keepaspectratio]{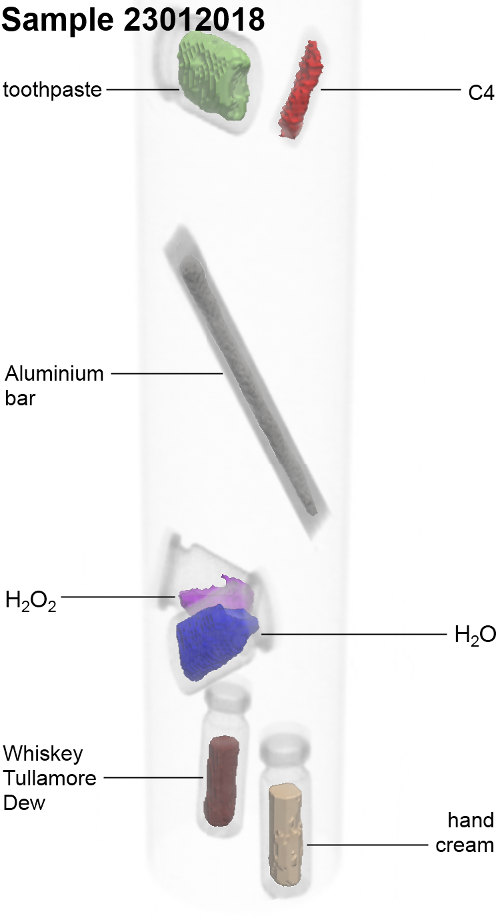}
  \caption{\label{fig:music3D:manSeg:23012018}Sample 23012018}
\end{center}
\end{figure}

\begin{figure}[htbp]
\begin{center}
  \centering
  \includegraphics[width=0.75\linewidth,height=12.1cm,keepaspectratio]{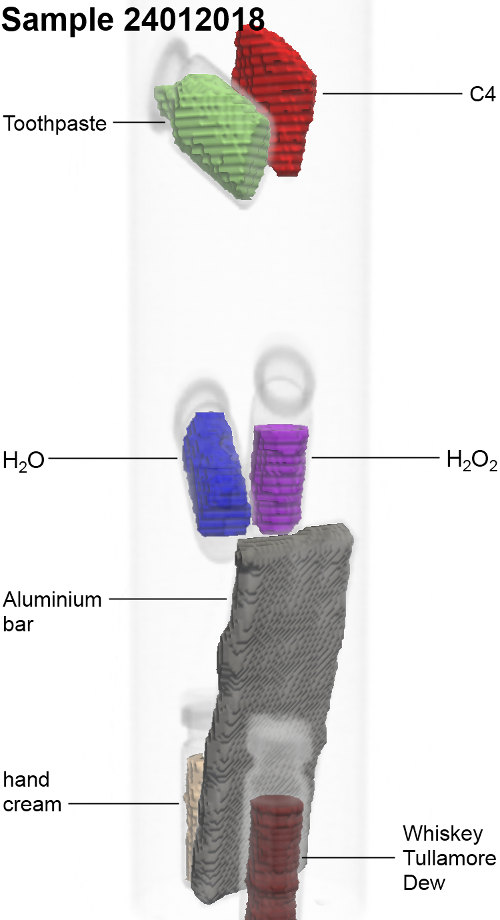}
  \caption{\label{fig:music3D:manSeg:24012018}Sample 24012018}
\end{center}
\end{figure}

\begin{figure}[htbp]
\begin{center}
  \centering
  \includegraphics[width=0.75\linewidth,height=12.1cm,keepaspectratio]{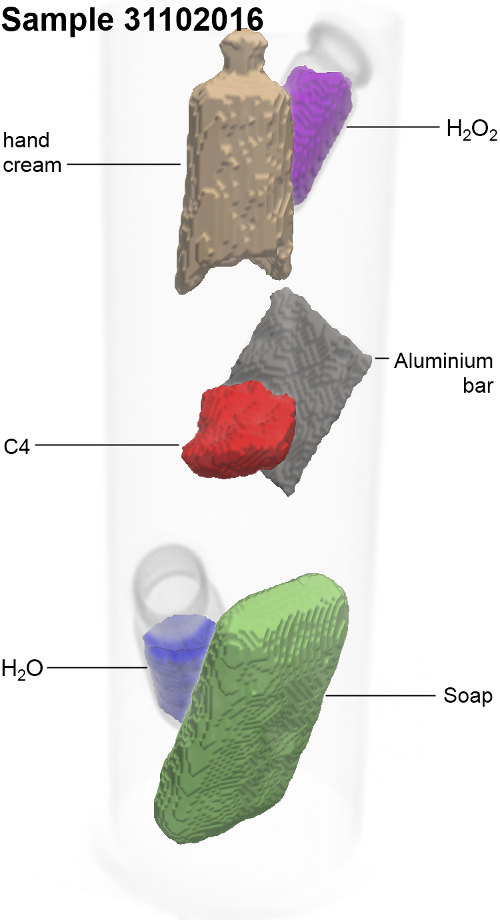}
  \caption{\label{fig:music3D:manSeg:31102016}Sample 31102016}
\end{center}
\end{figure}

\begin{figure}[htbp]
\begin{center}
  \centering
  \includegraphics[width=0.75\linewidth,height=12.1cm,keepaspectratio]{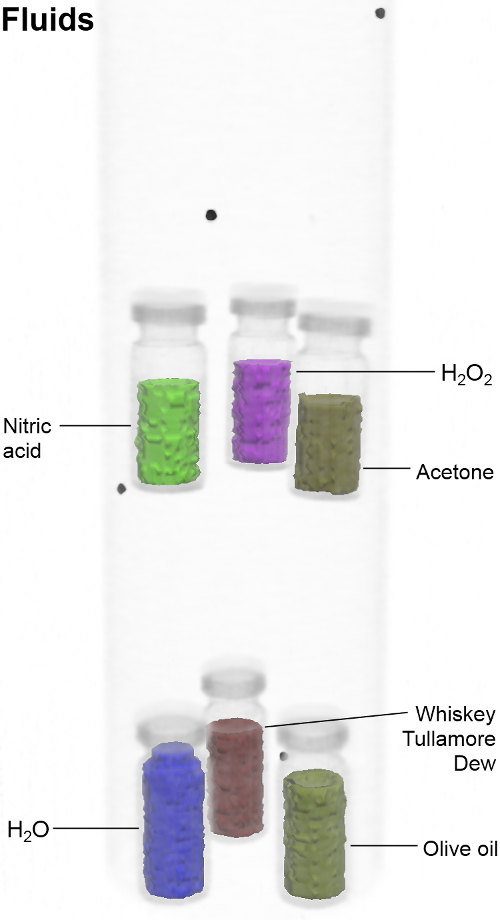}
  \caption{\label{fig:music3D:manSeg:Fluids}Fluids}
\end{center}
\end{figure}

\begin{figure}[htbp]
\begin{center}
  \centering
  \includegraphics[width=0.75\linewidth,height=12.1cm,keepaspectratio]{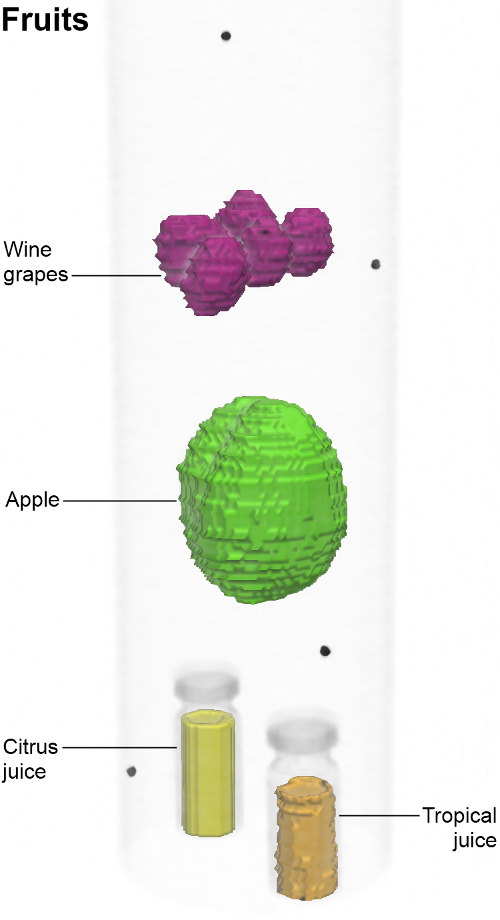}
  \caption{\label{fig:music3D:manSeg:Fruits}Fruits}
\end{center}
\end{figure}

\begin{figure}[htbp]
\begin{center}
  \centering
  \includegraphics[width=0.75\linewidth,height=12.1cm,keepaspectratio]{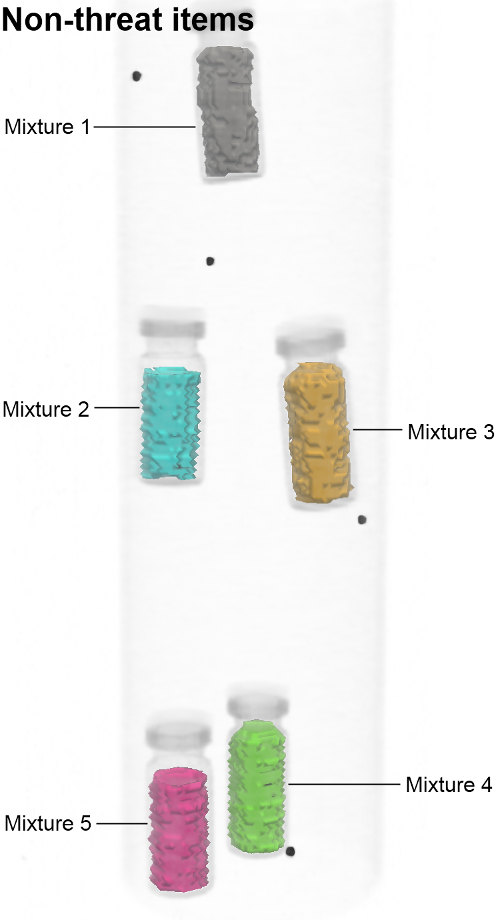}
  \caption{\label{fig:music3D:manSeg:NonThreat}Non-threat items}
\end{center}
\end{figure}

\begin{figure}[htbp]
\begin{center}
  \centering
  \includegraphics[width=0.75\linewidth,height=12.1cm,keepaspectratio]{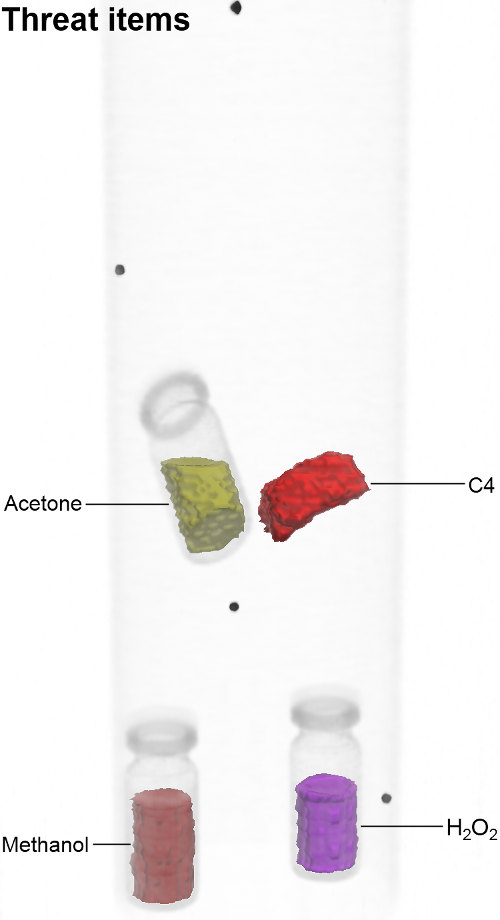}
  \caption{\label{fig:music3D:manSeg:Threat}Threat items}
\end{center}
\end{figure}

\end{document}